\documentclass{article}

\PassOptionsToPackage{numbers, compress}{natbib}

\usepackage[preprint]{neurips_2025}

\usepackage[utf8]{inputenc} 
\usepackage[T1]{fontenc}    
\usepackage{hyperref}       
\usepackage{url}            
\usepackage{booktabs}       
\usepackage{amsfonts}       
\usepackage{nicefrac}       
\usepackage{microtype}      
\usepackage[table]{xcolor}
\usepackage{amsmath}
\usepackage{wrapfig}
\usepackage{array}
\usepackage{tabularx}
\usepackage{arydshln}
\usepackage{makecell}

\newcolumntype{Y}{>{\centering\arraybackslash}X}

\usepackage{graphicx}
\usepackage{booktabs}
\usepackage{multirow}

\title{DAM: Domain-Aware Module \\ for Multi-Domain Dataset Condensation}

\author{
  Jaehyun Choi \quad
  Gyojin Han \quad
  Dong-Jae Lee \quad
  Sunghyun Baek \quad
  Junmo Kim \\
  Korea Advanced Institute of Science and Technology (KAIST) \\
  \texttt{\{chlwogus, hangj0820, jhtwosun, baeksh, junmo.kim\}@kaist.ac.kr}
}

\begin{document}
\maketitle
\begin{abstract}
Dataset Condensation (DC) has emerged as a promising solution to mitigate the computational and storage burdens associated with training deep learning models.
However, existing DC methods largely overlook the multi-domain nature of modern datasets, which are increasingly composed of heterogeneous images spanning multiple domains.
In this paper, we extend DC and introduce Multi-Domain Dataset Condensation (MDDC), which aims to condense data that generalizes across both single-domain and multi-domain settings.
To this end, we propose the Domain-Aware Module (DAM), a training-time module that embeds domain-related features into each synthetic image via learnable spatial masks.
As explicit domain labels are mostly unavailable in real-world datasets, we employ frequency-based pseudo-domain labeling, which leverages low-frequency amplitude statistics.
DAM is only active during the condensation process, thus preserving the same images per class (IPC) with prior methods.
Experiments show that DAM consistently improves in-domain, out-of-domain, and cross-architecture performance over baseline dataset condensation methods.
\end{abstract}
\vspace{-1ex}
\section{Introduction}
\label{sec:intro}
\vspace{-1ex}

Over the past decade, deep learning models have grown substantially in capacity, achieving remarkable progress in diverse tasks across vision, language, and multi-modal domains.
This performance growth has been tightly coupled with the dataset size.
To meet this demand, data collection has shifted from manual curation to automated web crawling, yielding datasets that are not only large in size but also highly heterogeneous.
These datasets often consist of samples drawn from various domains with drastically different visual characteristics, including changes in texture, lighting, color distribution, and abstraction level.
While such diversity benefits model robustness, it also raises new challenges for training efficiency and data quality management at scale.

Dataset Condensation (DC) has emerged as a promising direction to reduce training cost by synthesizing a small set of highly informative samples.
First introduced by Wang et al.~\cite{wang2018dataset}, DC replaces the original training data with a compact synthetic dataset, optimized to preserve the training dynamics of real data.
Recent DC methods improve this core idea using gradient matching~\cite{dc_2021}, distribution alignment~\cite{dm_2023}, or trajectory matching~\cite{mtt_2022}, and have shown strong results on curated benchmarks such as CIFAR~\cite{cifar_2009}.
These approaches significantly reduce training time and memory usage.
However, nearly all existing methods assume that the dataset is homogeneous in style.

\begin{wrapfigure}{r}{0.5\textwidth}
    \centering
    \includegraphics[width=0.48\textwidth]{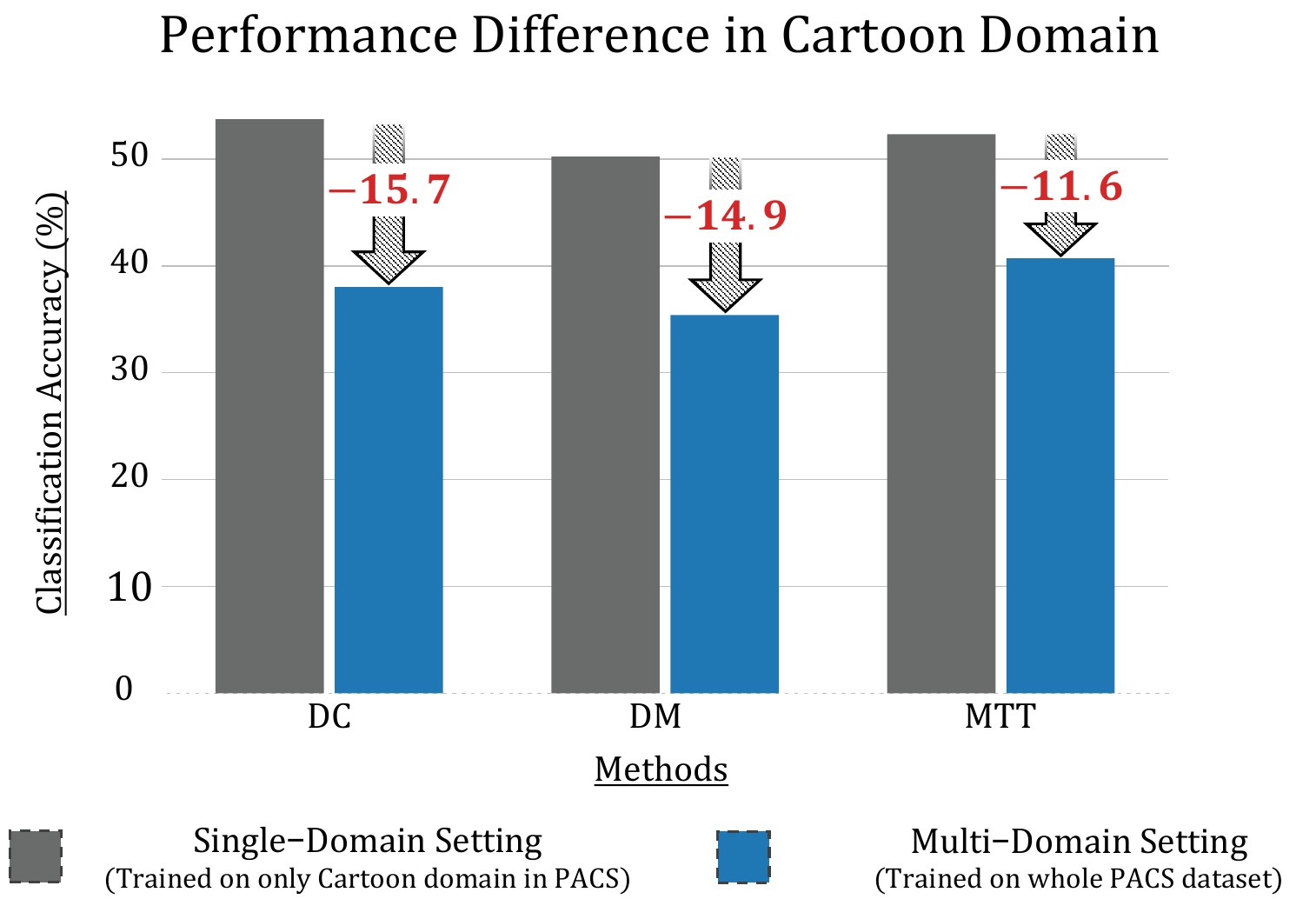}
    \caption{Performance of single- and multi-domain training for existing dataset condensation methods (DC, DM, MTT) on the PACS dataset under a 10 images per class setting. In the single-domain setup, models are trained using only Cartoon domain images, assuming access to explicit domain labels. In contrast, the multi-domain dataset setting trains on the full PACS dataset without domain supervision, reflecting modern datasets. In all prior methods, the performance drop in the multi-domain setting was significant.}
    \label{fig:intro}
\end{wrapfigure}

This assumption breaks down in many realistic scenarios, where datasets contain images from multiple visual domains (e.g., photo, logo drawing, advertisement, art-painting, etc.) due to their mixed-source construction.
When DC is applied to such mixed-domain data, synthetic images often collapse toward dominant domain styles, leading to degraded performance.
Figure~\ref{fig:intro} illustrates this problem on the PACS~\cite{pacs_iccv_2017} dataset, which includes 7 classes across 4 domains (e.g., Art-Painting, Cartoon, Photo, and Sketch).
We compare existing DC methods under two training settings, evaluating all models on the Cartoon domain.
In the \textbf{single-domain setting}, the Cartoon domain is isolated using oracle domain labels, resulting in higher accuracy, but this relies on metadata that is typically unavailable in a real-world setting.
In contrast, the \textbf{multi-domain setting} uses the full PACS dataset without domain separation, reflecting a more realistic scenario.
Here, all prior methods suffer significant performance drops, often exceeding 10\%p.
One might consider mitigating this by condensing data separately per domain or using ground-truth domain labels during condensation.
However, both strategies are fundamentally flawed within the DC framework.
Per-domain condensation scales linearly with the number of domains, inflating the synthetic data size, while relying on domain labels assumes costly and often unavailable annotations.

To fill this gap, we introduce the Multi-Domain Dataset Condensation (MDDC) task with the goal of synthesizing a single compact dataset that performs well across both single-domain and multi-domain settings, without explicit domain labels or increasing the Images per Class (IPC).
To solve this, we propose the Domain-Aware Module (DAM), a training-time module that equips each synthetic image with the capacity to represent multiple domain styles through learnable spatial masks.
DAM enables the model to assign domain-specific features to different regions within an image, effectively encoding domain variation without requiring per-domain images.
As explicit domain labels are unavailable, we propose a lightweight pseudo-domain labeling scheme based on frequency-domain characteristics.
Specifically, we assign pseudo-domain labels by sorting real images by the mean amplitude of their low-frequency FFT components, a heuristic inspired by its success in domain adaptation~\cite{fda, fft}.
These pseudo-labels are used to supervise DAM during training and are discarded afterward.
DAM does not increase the number of synthetic images and introduces no additional overhead when training downstream models.
Through comprehensive experiments on five datasets, including CIFAR-10, CIFAR-100, Tiny ImageNet, PACS, and VLCS, and across architectures, including ConvNet, VGG, and ViT, we show that prior methods' performance deteriorates in the multi-domain setting and that DAM consistently improves performance in both in-domain and cross-domain evaluation settings, without compromising the efficiency goals of dataset condensation.
\vspace{-1ex}
\section{Related Works}
\label{sec:relatedworks}
\vspace{-1ex}

\subsection{Dataset Condensation}

As machine learning models have become larger and more complex, the amount of data required for training those models has also grown significantly.
In this context, the emergence of massive datasets has greatly increased the burden on computational resources and training time, creating a bottleneck in model development.
Dataset distillation~\cite{wang2018dataset} is a formulation proposed to address this issue by compressing a large dataset into a much smaller synthetic dataset while still maintaining the essential data characteristics of the original dataset for training deep learning models.
This approach drastically reduces training time and computational costs, allowing models trained on the condensed dataset to achieve performance comparable to those trained on the original, large-scale datasets.
Among various strategies in dataset condensation, including gradient matching methods~\cite{wang2018dataset,dc_2021,kim2022dataset}, approaches based on distribution matching~\cite{dm_2023,zhao2023improved}, trajectory matching~\cite{mtt_2022,guo2024towards}, and generative-model-based approaches leveraging GANs or diffusion models~\cite{glad, d4m}, we focus on gradient matching, distribution matching, and trajectory matching.

\vspace{-2ex}

\paragraph{Gradient Matching} 
Dataset distillation methods based on gradient matching aim to match the gradients of a neural network that are calculated for a loss function over a synthetic dataset and the original dataset for the purpose of dataset condensation.
DC~\cite{dc_2021} first formulated the dataset distillation as a minimization problem between gradients that are calculated from an original dataset and a condensed dataset.
IDC~\cite{kim2022dataset} improved data condensation by efficiently parameterizing synthetic data to preserve essential characteristics with a smaller dataset, and they generated multiple formations of data to maintain model performance while significantly reducing storage and computation costs.
Zhang et al.~\cite{zhang2023accelerating} accelerated the distillation process by utilizing models in the early stages of training, rather than calculating gradients with randomly initialized models as in existing gradient-matching-based dataset distillation methods.
To address the resulting lack of model diversity, they introduced a model augmentation technique by adding small perturbations to the parameters of selected early-stage models.

\vspace{-2ex}

\paragraph{Distribution Matching}
Dataset distillation methods based on distribution matching were proposed to overcome the limitations of gradient matching methods, which require complex optimization and high computational costs.
DM~\cite{dm_2023} introduced a method that aligns the distribution of the original and synthetic datasets in embedding space, significantly improving the efficiency of dataset distillation and enabling condensed datasets to retain performance close to that of the original data, even with fewer data points.
IDM~\cite{zhao2023improved} enhanced distribution matching by addressing class imbalance and embedding issues.
They introduced new techniques, including split-and-augment augmentation, enhanced model sampling, and class-aware distribution normalization, to improve the diversity and representativeness of condensed datasets.

\vspace{-2ex}

\paragraph{Trajectory Matching}
MTT~\cite{mtt_2022} developed a method to create condensed datasets by mimicking the training trajectories of models trained on the original dataset.
By aligning the synthetic dataset's training path with that of the original data, they significantly improved the efficiency of dataset distillation.
FTD~\cite{du2023minimizing} improved trajectory matching by addressing the accumulated trajectory error, which often led to discrepancies between training and evaluation performance.
DATM~\cite {guo2024towards} addressed limitations in prior dataset distillation methods by introducing difficulty-aligned trajectory matching.
This approach enabled effective distillation without performance loss, even as the synthetic dataset size changes, and overcomes issues with prior methods' inability to adapt to different pattern difficulties.

\subsection{Domain-Aware Learning Approaches}

Research in domain-aware learning is crucial to addressing performance degradation caused by discrepancies between different domains.
Machine learning models tend to perform optimally when the distribution of training data matches that of test data. However, in real-world applications, data is often collected across various domains with distinct distributions.
These domain shifts can significantly impact a model’s generalization performance; without addressing these differences, models may only be effective in limited, specific environments.
Two prominent approaches to mitigate this issue are domain adaptation and domain generalization. Domain adaptation~\cite{ganin2016domain, long2018conditional, zhang2019bridging, zhu2023patch} focuses on improving the model's performance on a target domain by leveraging knowledge from a source domain where training data is available.
This typically involves techniques that reduce distributional differences between the source and target domains or map features from both domains onto a common representation. 
In contrast, domain generalization~\cite{li2018domain, zhou2021domain, cha2021swad, yao2022pcl, tcx} aims to build a model that can generalize to new, unseen domains without direct access to their data.
Domain generalization methods utilize multiple source domains to create a robust model that would perform equally well in various unseen domains.

Our plug-and-play method for multi-domain dataset condensation is related to previous domain-aware learning methods as it differentiates domains within the training dataset and considers possible domain shifts.
To the best of our knowledge, Domain-Aware Module (DAM) is the first work to incorporate domain-awareness into dataset condensation, bridging a previously unexplored gap between dataset condensation and multi-domain dataset.
\begin{figure*}[t]
    \centering
    \includegraphics[width=4.85in]{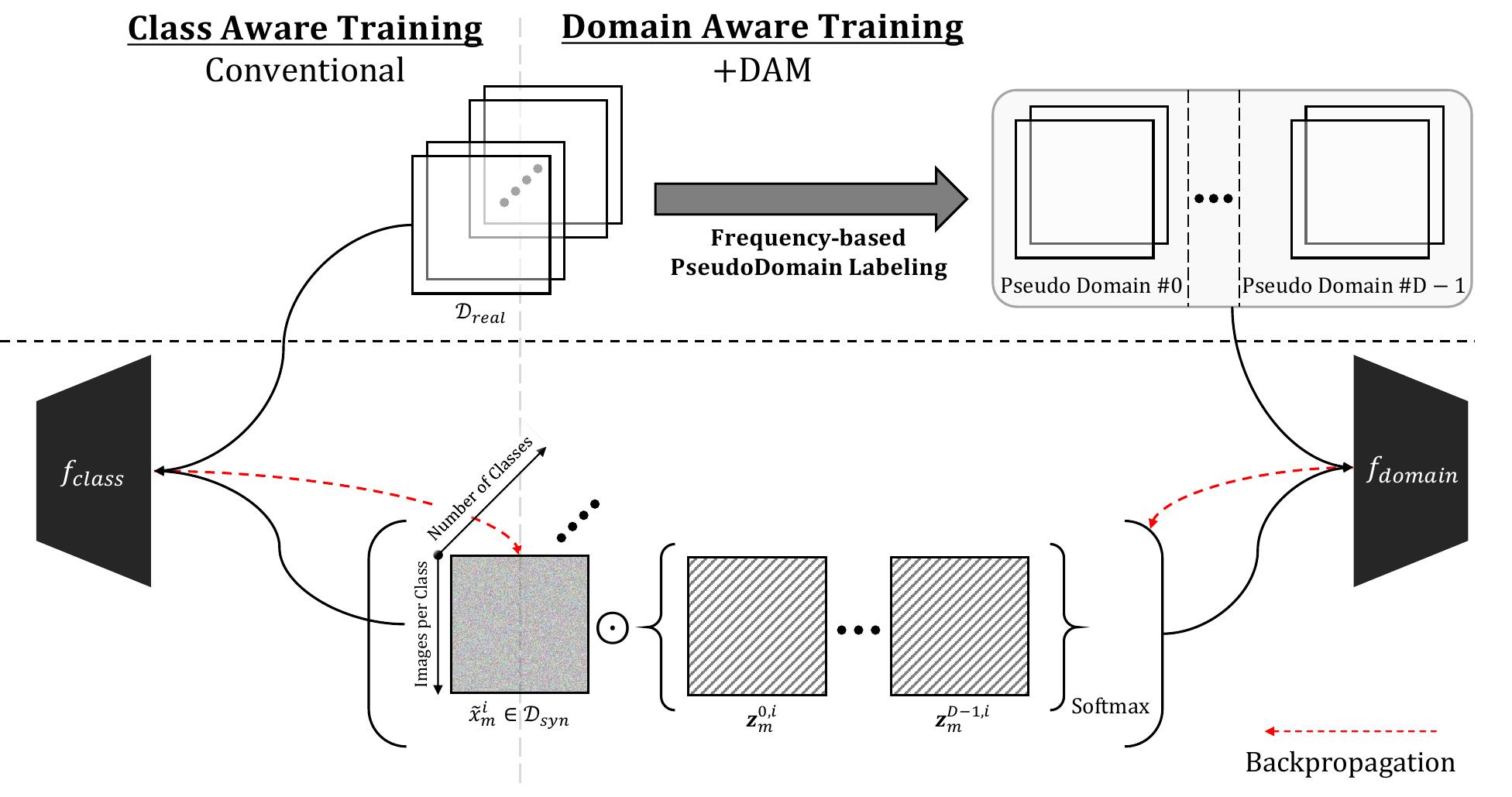}
    \caption{DAM incorporates both class-aware training (left) from prior methods and domain-aware training (right), the proposed DAM.}
    \label{fig:mainfig}
    \vspace{-1em}
\end{figure*}

\vspace{-1ex}
\section{Method}
\vspace{-1ex}

Given a dataset \(\mathcal{D}_{\text{real}}=\{x_n,y_n\}_{n=1}^{N}\) where \(y_n\in\{0,\cdots,C-1\}\), single-domain dataset condensation aims to synthesize a much smaller synthetic dataset \(\mathcal{D}_{\text{syn}}=\{\tilde{x}_m, \tilde{y}_m\}_{m=1}^M\) where \(M \ll N\) such that \(\mathcal{D}_{\text{syn}}\) has the same or similar power as \(\mathcal{D}_{\text{real}}\) in terms of model training.
In Multi-Domain Dataset Condensation (MDDC), it takes a step further and encodes domain variability within the \(\mathcal{D}_{\text{syn}}\) without explicit domain labels while preserving the class-discriminative features.

\subsection{Domain-Aware Module}

Each synthetic image \(\tilde{x}_m\in\mathbb{R}^{H \times W \times 3}\) is paired with a learnable domain mask \(\mathbf{z}_m^{d,i}\in\mathbb{R}^{H\times W\times 3}, d=\{0,\dots,D-1\}\), where \(D\) is the number of pseudo domains and \(i\) denotes the current iteration.
We initialize all elements in the domain mask with 0.01 at \(i=0\).
To prevent a single domain dominating the synthetic image, we leverage a per-pixel temperatured softmax to generate relative importance of each domain to each synthetic image, as well as to balance the domain importance among the \(\mathbf{z}_m^{d,i}\) as follows:
\begin{equation}
\alpha_m^{d,i} = \frac{\exp(\mathbf{z}_m^{d,i}/\tau)}{\sum_{d'=0}^{D-1} \exp(\mathbf{z}_m^{d',i}/\tau)},
\end{equation}
where \(\alpha_m^{d,i}\in\mathbb{R}^{H \times W \times 3}\) and \(\tau\) is the temperature factor in the softmax function.
Through \(\alpha_m^{d,i}\), a synthetic image saliency map with domain \(d\) at iteration \(i\) is obtained as
\begin{equation}
\tilde{x}_m^{d,i} = \tilde x_m^{i} \odot \alpha_m^{d,i},
\label{eq:eq2}
\end{equation}
where \(\odot\) is element-wise multiplication, and this satisfies the exact reconstruction identity as
\begin{equation}
    \tilde{x}_m^i = \sum_{d=0}^{D-1}\tilde x_m^{d,i},
\end{equation}
since \(\sum_{d=0}^{D-1}\alpha_m^{d, i}=1\) as it is output of softmax function.

Through a domain-aware module, several domains can coexist in disjoint spatial regions without
information loss.
Note that \(\mathbf{z}_m^{d,i}\) is trained along with \(\tilde{x}_m^i\) during training.
As both \(\mathbf{z}_m^{d,i}\) and \(\tilde{x}_m^{i}\) are trained, \(\tilde{x}_m^{d,i}\) are updated by Equation~\ref{eq:eq2}.

\subsection{Frequency-Based Pseudo Domain Labeling}

In many curated multi-domain benchmarks (e.g., PACS, Office-Home), explicit domain labels are available as domain differences are mostly distinguishable.
However, for unconstrained web data or large mixed datasets, explicit domain labels are mostly unavailable, primarily because the goal of the dataset is not for classifying the domains but also because the distinction for each domain is vague or overlapping.
We first define {\it{domain}} as variation not attuned with task-relevant information, in this case class-discriminative features, and leverage Fast Fourier Transformation (FFT) to extract domain-specific information for each image in \(\mathcal{D}_{\text{real}}\) as theoretically supported and applied in prior domain adaptation and domain generalization~\cite{fft, fda}.
For every real image \(x_n\) the discrete 2-D Fourier
transform as follows
\begin{equation}
\mathcal{F}(x_n)[u,v]=\sum_{h=0}^{H-1}\sum_{w=0}^{W-1} x_n[h,w]\, e^{-j2\pi(uh/H+vw/W)},
\end{equation}
which is computed per color channel.
Through shifting, the center of the amplitude becomes the low-frequency region, which prior domain adaptation and domain generalization methods leveraged for domain-specific information.
Likewise, we crop the central region with a cropping ratio \(\beta\) and get the mean of the amplitude, \(\mu_n\), as follows:
\begin{equation}
    \mu_n = \frac{1}{3\beta^2HW}(Crop_{\beta}\{|\mathcal{F}(x_n)|_{\text{shifted}}\}\in\mathbb{R}^{\beta H \times \beta W \times 3}).
\end{equation}
Sorting \(\{\mu_n\}_{n=1}^{N}\) in ascending or descending order and slicing it into
\(D\) equal parts assigns pseudo-domain labels
\[
d_n=
\Bigl\lfloor\tfrac{\mbox{ranking}(\mu_n)-1}{N/D}\Bigr\rfloor,
\qquad
d_n\in\{0,\dots,D-1\}, \mbox{ranking}(\mu_n)\in\{1,...N\}.
\]

\subsection{Training objective}
Being a plug-and-play module, we leverage the same loss function from the prior base models \(\mathcal{L}_{base}\) and the loss for the class becomes
\begin{equation}
    \mathcal{L}_{\text{cls}}=\mathcal{L}_{base}(\Theta; \mathcal{D}_{\text{real}}, \mathcal{D}_{\text{syn}}),
\end{equation}
where \(\Theta\) is the parameters needed for the loss computation.
Accordingly, we define the domain loss as
\begin{equation}
    \mathcal{L}_{\text{dom}}=\mathcal{L}_{base}(\Theta'; \mathcal{D}_{\text{real}}, \mathcal{D}_{\text{syn}}^{\text{dom}}).
\end{equation}
Here, parameters for the domain loss are denoted as \(\Theta'\).
The \(\mathcal{D}^{\text{dom}}_{\text{syn}} = \left\{ \tilde{\mathbf{x}}_m^d, \tilde{y}_m \right\}_{1 \leq m \leq M, 0 \leq d < D}\) where \(\tilde{x}_m^d=\tilde{x}_m \odot \alpha_m^d\), is used solely to supervise the Domain-Aware Module (DAM) during the condensation phase.
\textbf{Notably, the domain masks \(\mathbf{z}_m^{d,i}\) used to compute \(\alpha_m^d\) are discarded after condensation.}
As a result, only \(\tilde{x}_m\in\mathcal{D}_{syn}\) is used during downstream training, ensuring that the number of synthetic images remains unchanged and the Images per Class is preserved.

The architecture for domain loss is identical to class loss, but the parameters are initialized differently.
Also, note that the \(\mathcal{D}_{real}\) for class and domain loss is the same while the batch configuration differs.
For the class loss, the batch is grouped by the class label following the prior methods, while it is grouped by the pseudo-domain label for the domain loss.
To sum up, the final loss becomes
\begin{equation}
    \mathcal{L}_{\text{total}}=\mathcal{L}_{\text{cls}}+\lambda\mathcal{L}_{\text{dom}},
\end{equation}
where \(\lambda\) is the weighting factor.
\(\mathcal{L}_{\text{cls}}\) provides gradients for updating \(\tilde{x}_m^i\) and \(\mathcal{L}_{\text{dom}}\) for updating both \(\tilde{x}_m^i\) and \(\mathbf{z}_m^{d,i}\).
Parameters \(\Theta\) and \(\Theta'\) are randomly initialized or frozen after being trained on real data, depending on the prior method (Details are given in the supplementary material~\ref{supple:theta_details}).
The overall pipeline is illustrated in Figure~\ref{fig:mainfig}.
\vspace{-1ex}
\section{Experiments}
\label{sec:exp}
\vspace{-1ex}

\subsection{Dataset}

We evaluate our plug-and-play method, DAM, on \(32\times32\) CIFAR-10 and CIFAR-100~\cite{cifar_2009}, and \(64\times64\) Tiny ImageNet~\cite{tiny_2015}, the three most commonly used datasets in the field of dataset condensation.
The experiment setting with these datasets is \textbf{single-domain setting}.
Additionally, we employ \(64\times64\) PACS~\cite{pacs_iccv_2017}, VLCS~\cite{vlcs_2013}, and Office-Home~\cite{oh_cvpr_2017} datasets that are commonly used in the field of domain adaptation (DA) and domain generalization (DG).
These multi-domain datasets have four distinct domains and are leveraged not only to validate the effectiveness of DAM in \textbf{multi-domain setting} but also to better analyze the differences between single- and multi-domain dataset settings.
We note that the provided domain labels are not leveraged unless explicitly stated in the experiment setting.

\subsection{Implementation Details}

We implement DAM on three pioneering prior methods, DC~\cite{dc_2021}, DM~\cite{dm_2023}, and MTT~\cite{mtt_2022}, in gradient matching, distribution matching, and trajectory matching dataset condensation, respectively.
For a fair comparison, we follow the conventional experiment settings employing ConvNet architecture~\cite{convnet_2018} while varying the depth of the network depending on the image size of the \(D_{real}\).
More specifically, three-depth ConvNet is utilized for all experiments with the CIFAR-10 and CIFAR-100 datasets, while all the other datasets leverage four-depth ConvNet.
All of the hyperparameters introduced in each prior method are set identically, and the learning rate for the DG datasets is set equal to the Tiny ImageNet setting.
We note that \(D_{syn}\) is initialized with \textbf{Gaussian noise} in all of our experiments rather than initializing with \textbf{real} image from \(D_{real}\) as some prior works do.
However, we demonstrate that the performance gap still persists even when initializing with \text{real} image in the supplementary material~\ref{supple:real_initial}.
\(\mathbf{z}_m^{d,i}\) is initialized with 0.01 and the temperature \(\tau\) is set to 0.1 when applying softmax among the \(D\) masks for all experiments.
The domain embedding weight \(\lambda\) is set to 0.1 for DC and DM and 0.01 for the MTT.
All of the hyperparameter sweep experiments (e.g., \(\mathbf{z}_m\) initial value, domain embedding weight \(\lambda\), and temperature \(\tau\) value) can be found in the supplementary material~\ref{supple:hyperparameter_sweep}.
\(D\) is set to 4 for all experiments.
Finally, we follow DM~\cite{dm_2023} for the evaluation protocol for all the experiments, and the results presented in the tables are the average of 10 evaluation results.

\begin{table}[t!]
\caption{Results with and without DAM on the prior methods on the \textbf{single-domain} setting. ``T.Image.'' denotes Tiny ImageNet dataset. All results are the average of 10 runs and reported as mean \(\pm\) standard deviation.}
\label{tab:single_domain}
\renewcommand{\arraystretch}{0.95}
\setlength{\tabcolsep}{4pt}
\centering
\small
\begin{tabularx}{\textwidth}{lYYYYYYY}
\toprule
\textbf{Dataset} & \multicolumn{3}{c}{\textbf{CIFAR-10}} & \multicolumn{3}{c}{\textbf{CIFAR-100}} & \textbf{T.Image.} \\
\cmidrule(lr){1-1} \cmidrule(lr){2-4} \cmidrule(lr){5-7} \cmidrule(lr){8-8}
Img/Cls & 1 & 10 & 50 & 1 & 10 & 50 & 1 \\
Ratio (\%) & 0.02 & 0.2 & 1 & 0.2 & 2 & 10 & 0.2 \\
\midrule
\midrule
Random & 12.5\(_{\pm 0.8}\) & 25.1\(_{\pm 1.4}\) & 42.5\(_{\pm 0.5}\) & 3.7\(_{\pm 0.2}\) & 13.9\(_{\pm 0.3}\) & 29.0\(_{\pm 0.3}\) & 1.3\(_{\pm 0.1}\) \\
\midrule
DC & 27.4\(_{\pm 0.2}\) & 43.3\(_{\pm 0.3}\) & 53.0\(_{\pm 0.3}\) & 12.2\(_{\pm 0.3}\) & 24.8\(_{\pm 0.3}\) & - & - \\
\textbf{DC + DAM} & \textbf{29.0\(_{\pm 0.5}\)} & \textbf{45.4\(_{\pm 0.3}\)} & \textbf{54.5\(_{\pm 0.2}\)} & \textbf{13.0\(_{\pm 0.2}\)} & \textbf{25.8\(_{\pm 0.1}\)} & - & - \\
[0.1ex] \cdashline{1-8}[3pt/3pt] \\ [-1.8ex]
DM & 24.7\(_{\pm 0.3}\) & 47.4\(_{\pm 0.4}\) & 58.2\(_{\pm 0.1}\) & 10.9\(_{\pm 0.2}\) & 29.2\(_{\pm 0.2}\) & 36.5\(_{\pm 0.2}\) & 3.7\(_{\pm 0.1}\) \\
\textbf{DM + DAM} & \textbf{27.1\(_{\pm 0.3}\)} & \textbf{49.8\(_{\pm 0.5}\)} & \textbf{59.5\(_{\pm 0.2}\)} & \textbf{11.8\(_{\pm 0.2}\)} & \textbf{30.0\(_{\pm 0.1}\)} & \textbf{37.3\(_{\pm 0.2}\)} & \textbf{4.2\(_{\pm 0.1}\)} \\
[0.1ex] \cdashline{1-8}[3pt/3pt] \\ [-1.8ex]
MTT & 41.9\(_{\pm 0.4}\) & 50.7\(_{\pm 0.8}\) & - & 15.8\(_{\pm 0.3}\) & 35.3\(_{\pm 0.2}\) & - & 4.8\(_{\pm 0.3}\) \\
\textbf{MTT + DAM} & \textbf{46.8\(_{\pm 0.4}\)} & \textbf{57.9\(_{\pm 0.4}\)} & - & \textbf{24.0\(_{\pm 0.3}\)} & \textbf{35.9\(_{\pm 0.2}\)} & - & \textbf{5.7\(_{\pm 0.2}\)} \\
\midrule
Whole Dataset & \multicolumn{3}{c|}{84.8\(_{\pm 0.1}\)} & \multicolumn{3}{c|}{56.2\(_{\pm 0.3}\)} & 37.6\(_{\pm 0.4}\) \\
\bottomrule
\end{tabularx}
\end{table}

\begin{table}[t!]
\caption{Results with and without DAM on the prior methods on the \textbf{multi-domain} setting. All results are the average of 10 runs and reported as mean \(\pm\) standard deviation.}
\label{tab:multi_domain}
\renewcommand{\arraystretch}{0.95}
\centering
\small
\begin{tabularx}{\textwidth}{lYYYYYY}
\toprule
\textbf{Dataset} & \multicolumn{2}{c}{\textbf{PACS}} & \multicolumn{2}{c}{\textbf{VLCS}} & \multicolumn{2}{c}{\textbf{Office-Home}} \\
\cmidrule(lr){1-1} \cmidrule(lr){2-3} \cmidrule(lr){4-5} \cmidrule(lr){6-7}
Img/Cls & 1 & 10 & 1 & 10 & 1 & 10 \\
Ratio (\%) & 0.08 & 0.8 & 0.07 & 0.7 & 0.46 & 4.6 \\
\midrule
\midrule
Random & 18.1\(_{\pm 2.6}\) & 33.0\(_{\pm 0.8}\) & 17.3\(_{\pm 2.2}\) & 27.0\(_{\pm 1.6}\) & 3.9\(_{\pm 0.3}\) & 12.9\(_{\pm 0.6}\) \\
\midrule
DC & 35.3\(_{\pm 0.6}\) & 46.1\(_{\pm 0.7}\) & 29.6\(_{\pm 0.9}\) & 39.0\(_{\pm 0.6}\) & 11.0\(_{\pm 0.3}\) & - \\
\textbf{DC + DAM} & \textbf{38.8\(_{\pm 0.7}\)} & \textbf{48.3\(_{\pm 0.5}\)} & \textbf{34.8\(_{\pm 1.0}\)} & \textbf{42.7\(_{\pm 0.5}\)} & \textbf{12.4\({\pm 0.4}\)} & - \\
[0.1ex] \cdashline{1-7}[3pt/3pt] \\ [-1.8ex]
DM & 28.7\({\pm 0.5}\) & 46.7\({\pm 0.5}\) & 29.1\({\pm 1.7}\) & 42.0\({\pm 0.3}\) & 9.0\({\pm 0.3}\) & 25.5\({\pm 0.3}\) \\
\textbf{DM + DAM} & \textbf{34.7\({\pm 1.1}\)} & \textbf{50.9\({\pm 0.4}\)} & \textbf{36.7\({\pm 1.1}\)} & \textbf{44.4\({\pm 0.3}\)} & \textbf{10.4\({\pm 0.3}\)} & \textbf{27.2\({\pm 0.3}\)} \\
[0.1ex] \cdashline{1-7}[3pt/3pt] \\ [-1.8ex]
MTT & 39.7\({\pm 0.6}\) & 45.9\({\pm 0.8}\) & 28.5\({\pm 2.1}\) & - & 13.8\({\pm 0.2}\) & - \\
\textbf{MTT + DAM} & \textbf{46.6\({\pm 0.9}\)} & \textbf{50.6\({\pm 0.6}\)} & \textbf{39.7\({\pm 1.8}\)} & - & \textbf{16.3\({\pm 0.2}\)} & - \\
\midrule
Whole Dataset & \multicolumn{2}{c|}{72.0\({\pm 0.8}\)} & \multicolumn{2}{c|}{60.8\({\pm 0.6}\)} & \multicolumn{2}{c}{50.4\({\pm 0.8}\)} \\
\bottomrule
\end{tabularx}
\end{table}

\subsection{Results}

\vspace{-1ex}
\paragraph{Main results}
Table~\ref{tab:single_domain} demonstrates the performance on the single-domain setting with three commonly utilized benchmarks in dataset condensation by varying Image per Class (IPC) and prior dataset condensation methods employed along with DAM.
Similarly, in Table~\ref{tab:multi_domain}, we showcase the performance on three commonly used benchmarks in DA and DG for the multi-domain setting.
Note, we use all of the domains in the dataset for training and evaluating the DG datasets.
The ``-'' in the tables denotes the experiment setting, which either 1) was not done in the original paper or 2) requires extensive computational resources beyond our limit.
All reported experiments show performance improvements when leveraging DAM with the prior methods.
This consistent improvement confirms that DAM effectively enriches condensed data with domain-specific structure while preserving class-discriminative information, the core objective in classification task.
Despite the risk that embedding additional domain cues might interfere with class semantics, the observed gains demonstrate that DAM successfully integrates domain context in a way that reinforces, rather than disrupts, the underlying class structure as intended.
\begin{figure}[t!]
    \centering
    \includegraphics[width=0.85\textwidth]{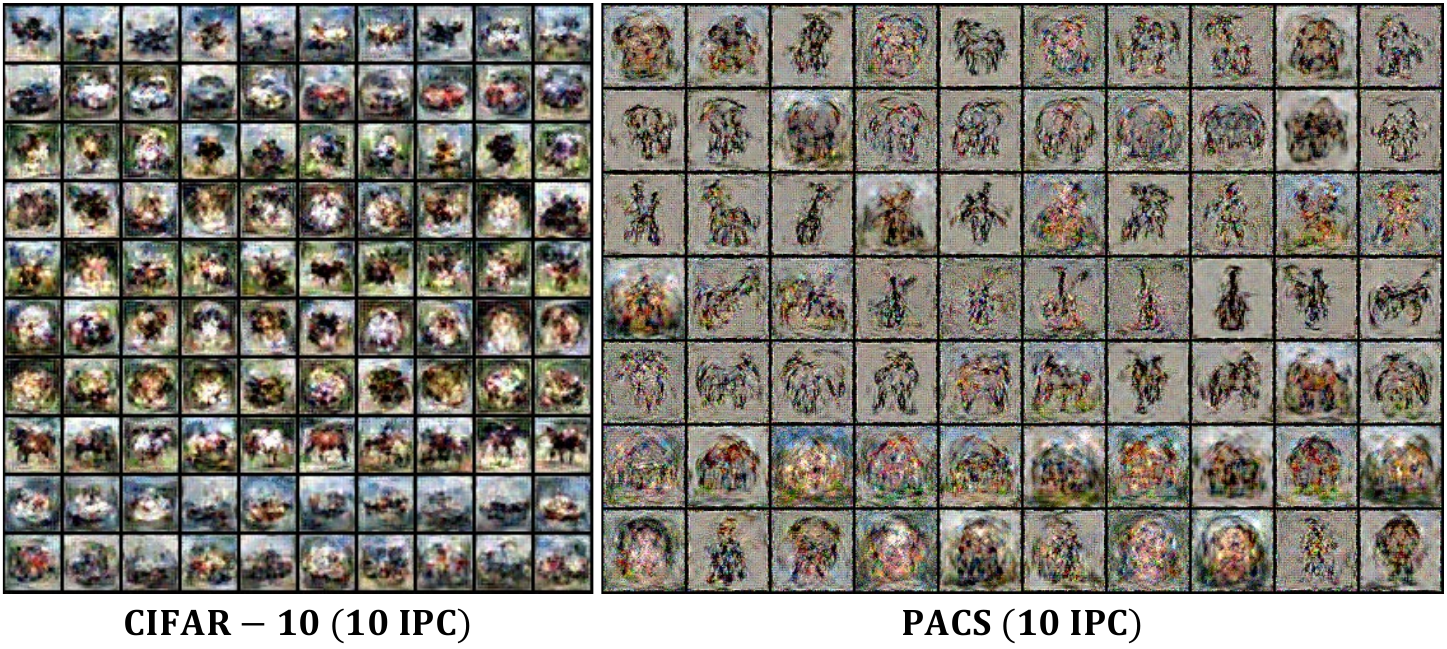}
    \caption{Visualization of the final output in CIFAR-10 and PACS under 10 IPC setting. The shown images are condensed with DC+DAM. More outputs can be found in the supplementary material.}
    \label{fig:qualitative_results}
\end{figure}
We visualize the final condensed synthetic data for CIFAR-10 and PACS datasets under 10 IPC setting on DC+DAM in Figure~\ref{fig:qualitative_results}.
Note that \(D_\text{syn}^\text{dom}\) is used only during dataset condensation process to generate \(D_\text{syn}\) and that all the results with DAM in Tables~\ref{tab:single_domain}, \ref{tab:multi_domain}, and \ref{tab:cross_architecture} are obtained when the model is trained with only the dataset \(D_\text{syn}\) that has \(M\) datapoints i.e. without using \(D_\text{sym}^\text{dom}\) at all. 

\vspace{-2ex}

\paragraph{Cross-architecture generalization}

\begin{table}[t!]
\renewcommand{\arraystretch}{0.95}
\setlength{\tabcolsep}{4pt}
\centering
\small
\caption{Cross-architecture results with condensed CIFAR-10 data under 10 IPC with ConvNet on various architectures. All results are the average of 10 runs and reported as mean \(\pm\) standard deviation.}
\label{tab:cross_architecture}
\rowcolors{2}{gray!15}{white}
\begin{tabularx}{\textwidth}{lYYYYYY}
\toprule
\rowcolor{white}
\textbf{Method} & \textbf{ConvNet} & \textbf{AlexNet} & \textbf{VGG} & \textbf{ResNet18} & \textbf{ViT-Tiny} & \textbf{ViT-Small} \\
\midrule
DC & 43.3\({\pm0.3}\) & 15.0\({\pm3.4}\) & 34.6\({\pm0.2}\) & 18.7\({\pm0.5}\) & 21.7\({\pm0.6}\) & 21.7\({\pm0.5}\) \\
\textbf{DC + DAM} & \textbf{45.4\({\pm0.3}\)} & \textbf{22.8\({\pm1.2}\)} & \textbf{35.9\({\pm0.4}\)} & \textbf{19.5\({\pm0.6}\)} & \textbf{22.4\({\pm0.4}\)} & \textbf{22.4\({\pm0.4}\)} \\
[0.1ex] \cdashline{1-7}[3pt/3pt] \\ [-1.8ex]
DM & 47.4\({\pm0.4}\) & 36.1\({\pm0.4}\) & 39.9\({\pm0.3}\) & 36.9\({\pm0.8}\) & 26.6\({\pm0.5}\) & 27.1\({\pm0.5}\) \\
\textbf{DM + DAM} & \textbf{49.8\({\pm0.5}\)} & \textbf{39.0\({\pm0.3}\)} & \textbf{40.9\({\pm0.7}\)} & \textbf{39.8\({\pm1.0}\)} & \textbf{26.9\({\pm0.5}\)} & \textbf{27.4\({\pm0.4}\)} \\
[0.1ex] \cdashline{1-7}[3pt/3pt] \\ [-1.8ex]
MTT & 50.7\({\pm0.8}\) & 23.2\({\pm1.3}\) & 45.7\({\pm0.8}\) & 38.9\({\pm0.8}\) & 20.3\({\pm1.4}\) & 22.5\({\pm0.7}\) \\
\textbf{MTT + DAM} & \textbf{57.9\({\pm0.4}\)} & \textbf{24.0\({\pm1.0}\)} & \textbf{46.6\({\pm0.9}\)} & \textbf{41.1\({\pm0.7}\)} & \textbf{20.5\({\pm0.8}\)} & \textbf{22.8\({\pm1.0}\)} \\
\bottomrule
\end{tabularx}
\vspace{-1em}
\end{table}
We assess the generalization capabilities of condensed synthetic data across different architecture frameworks.
Following MTT~\cite{mtt_2022}, we experiment on ConvNet, AlexNet, VGG11, and ResNet-18.
Furthermore, we experiment on ViT-Tiny and ViT-Small, which prior methods did not experiment on.
The cross-architecture experiments are conducted with condensed CIFAR-10 data under 10 IPC with ConvNet.
As shown in Table~\ref{tab:cross_architecture}, incorporating DAM shows superior generalization performance across methods and architectures compared to those without DAM, demonstrating the robustness across architecture.
\vspace{-1ex}
\section{Discussion}
\label{sec:discussion}

\vspace{-1ex}
\paragraph{Single-domain and multi-domain dataset}

\begin{table}[t!]
    \renewcommand{\arraystretch}{0.95}
    \centering
    \caption{Experiment results of single- and multi-domain dataset settings. In a single-domain dataset setting, the target data is used during the condensation process, whereas in a multi-domain dataset setting, the whole PACS dataset is utilized. The evaluation is done in the target domain images for both settings. The value inside the parentheses denotes the difference between Multi-Domain with DAM and without DAM.}
    \label{tab:single_multi}
    \small
    \begin{tabularx}{\columnwidth}{c|l|YYYY}
        \toprule
        \multicolumn{2}{c|}{Method (\(\downarrow\)) / Target Domain (\(\rightarrow)\)} & Photo & Art-Painting & Cartoon & Sketch \\
        \midrule
        \midrule
        Single-Domain & DC & 50.6 & 29.6 & 53.7 & 43.8 \\
        [0.1ex] \cdashline{1-6}[3pt/3pt] \\ [-1.8ex]
        \multirow{2}{*}{Multi-Domain} & DC & 48.1 & 27.6 & 38.0 & 32.1 \\
        & \textbf{DC + DAM} & 49.4 \textcolor{blue}{(+1.3)} & 30.4 \textcolor{blue}{(+2.8)} & 40.4 \textcolor{blue}{(+2.4)} & 37.4 \textcolor{blue}{(+5.3)} \\
        \midrule
        Single-Domain & DM & 50.7 & 29.5 & 50.2 & 35.4 \\
        [0.1ex] \cdashline{1-6}[3pt/3pt] \\ [-1.8ex]
        \multirow{2}{*}{Multi-Domain} & DM & 46.8 & 21.3 & 35.4 & 22.6 \\
        & \textbf{DM + DAM} & 47.4 \textcolor{blue}{(+0.6)} & 29.3 \textcolor{blue}{(+8.0)} & 37.1 \textcolor{blue}{(+1.7)} & 30.9 \textcolor{blue}{(+8.3)} \\
        \midrule
        Single-Domain & MTT & 55.2 & 31.9 & 55.1 & 42.3 \\
        [0.1ex] \cdashline{1-6}[3pt/3pt] \\ [-1.8ex]
        \multirow{2}{*}{Multi-Domain} & MTT & 50.7 & 24.5 & 40.8 & 44.6 \\
        & \textbf{MTT + DAM} & 52.1 \textcolor{blue}{(+1.4)} & 26.9 \textcolor{blue}{(+2.4)} & 50.5 \textcolor{blue}{(+9.7)} & 50.9 \textcolor{blue}{(+6.3)} \\
        \bottomrule
    \end{tabularx}
\end{table}

The need for Multi-Domain Dataset Condensation (MDDC) methods has been highlighted in Section~\ref{sec:intro} as a figure.
We further extend the experiment on the same setting and show the results in Table~\ref{tab:single_multi}.
For a single-domain dataset setting, we isolate the target domain with an explicit domain label for condensing and evaluating.
On the other hand, for a multi-domain dataset setting, the whole PACS dataset (i.e., all four domains) is utilized for the training.
The evaluation was done on the same target domains for both single- and multi-domain dataset settings.
In most of the results, the single-domain dataset setting performed much better than the multi-domain dataset settings, demonstrating the need for multi-domain dataset consideration in DC.
Notably, each prior pioneering method with the DAM always performed better than without DAM, and in cases such as DC and DM with the Art-Painting as the target domain, the performance was on-par with the single-domain dataset setting.
Most importantly, the performance gap between the single-domain setting and the multi-domain setting substantially declined with DAM.

\vspace{-2ex}

\paragraph{Leave-one-domain-out evaluation}

\begin{table}[t!]
    \renewcommand{\arraystretch}{0.95}
    \centering
    \caption{Leave-one-domain-out evaluation on PACS, VLCS, and Office-Home datasets with 1 IPC using DM and DM+DAM. Target domains are abbreviated as: PACS — (P)hoto, (A)rt-Painting, (C)artoon, (S)ketch; VLCS — Pascal (V)OC, (L)abelMe, (C)altech, (S)un; Office-Home — (A)rt, (C)lipart, (P)roduct, (R)eal-World.}
    \label{tab:dg}
    \small
    \renewcommand{\arraystretch}{1.05}
    \setlength{\tabcolsep}{3pt}
    \begin{tabularx}{\textwidth}{l*{12}{Y}}
        \toprule
        Dataset
        & \multicolumn{4}{c}{PACS} 
        & \multicolumn{4}{c}{VLCS} 
        & \multicolumn{4}{c}{Office-Home} \\
        \cmidrule(lr){2-5} \cmidrule(lr){6-9} \cmidrule(lr){10-13}
        Target Domain & P & A & C & S & V & L & C & S & A & C & P & R \\
        \midrule
        DM       & 29.9 & 18.9 & 20.6 & 22.5 & 24.6 & 33.9 & 21.8 & 33.0 & 3.3 & 6.3 & 7.0 & 5.8 \\
        \textbf{DM+DAM}   & \textbf{44.4} & \textbf{24.4} & \textbf{27.7} & \textbf{30.7} 
                 & \textbf{26.9} & \textbf{40.0} & \textbf{26.0} & \textbf{36.3} 
                 & \textbf{5.2} & \textbf{7.4} & \textbf{9.7} & \textbf{7.1} \\
        \bottomrule
    \end{tabularx}
\end{table}
In this section, we evaluate whether embedding domain information into each synthetic image improves generalization to unseen domains beyond the training set.
For the experiment, we tested on the three domain generalization benchmarks and compared DM with DM+DAM under 1 IPC using explicit domain labels only to isolate the target domain, which is only used during evaluation and neglected during the condensation process.
As can be seen from Table~\ref{tab:dg}, employing DAM with DM performed better with a substantial gap.
This validates that employing DAM substantially increases the generalization ability of the condensed data through embedding informative and non-overlapping domain information.
These results showcase the possibility of using DAM even for domain adaptation and domain generalization, where the burden of gathering data is much more costly.

\vspace{-2ex}

\paragraph{Various pseudo-domain labeling}

\begin{table}[t!]
\centering
\small
\caption{Comparison of different pseudo-domain labeling strategies on the CIFAR-10, PACS, and VLCS datasets under 1 and 10 IPC. All results are the average of 10 runs and reported as mean \(\pm\) standard deviation. FFT: frequency feature extraction; 
log-Var: log-variance of early features; Mean-Sort: ordering features by mean value; K-Means: clustering features with K-Means. Baselines include random pseudo-labels and actual domain labels.}
\resizebox{\textwidth}{!}{
\begin{tabular}{cccccccccc}
\toprule
\multicolumn{4}{c|}{Pseudo Domain Labeling Method} & \multicolumn{6}{c}{Method} \\
\cmidrule(r){1-4} \cmidrule(l){5-10}
\multirow{2}{*}{FFT} & \multirow{2}{*}{log-Var} & \multirow{2}{*}{Mean-Sort} & \multirow{2}{*}{K-Means} & \multicolumn{2}{c}{CIFAR-10} & \multicolumn{2}{c}{PACS} & \multicolumn{2}{c}{VLCS} \\
& & & & 1 IPC & 10 IPC & 1 IPC & 10 IPC & 1 IPC & 10 IPC \\
\midrule
\checkmark &          & \checkmark &          & \textbf{27.1\({\pm0.3}\)} & \textbf{49.8\({\pm0.5}\)} & \textbf{34.7\({\pm1.1}\)} & \textbf{50.9\({\pm0.4}\)} & 36.7\({\pm1.1}\) & \textbf{44.4\({\pm0.3}\)} \\
\checkmark &          &           & \checkmark & 26.6\({\pm0.4}\) & 49.7\({\pm0.3}\) & 32.3\({\pm0.6}\) & 49.7\({\pm0.7}\) & \textbf{36.8\({\pm1.1}\)} & 44.3\({\pm0.4}\) \\
           & \checkmark & \checkmark &          & 27.0\({\pm0.3}\) & \textbf{49.8\({\pm0.2}\)} & 33.6\({\pm0.8}\) & 49.7\({\pm0.4}\) & 34.0\({\pm0.8}\) & 44.1\({\pm0.4}\) \\
           & \checkmark &           & \checkmark & 26.5\({\pm0.4}\) & 49.4\({\pm0.3}\) & 33.5\({\pm0.6}\) & 48.8\({\pm0.6}\) & 35.3\({\pm1.1}\) & 44.0\({\pm0.4}\) \\
\midrule
\multicolumn{4}{c}{Random Pseudo Labels} & 25.3\({\pm0.5}\) & 48.1\({\pm0.7}\) & 31.7\({\pm1.4}\) & 48.0\({\pm1.2}\) & 31.1\({\pm1.8}\) & 42.4\({\pm0.7}\) \\
\multicolumn{4}{c}{Actual Domain Labels} & - & - & 34.0\({\pm1.7}\) & 50.6\({\pm0.6}\) & 34.5\({\pm1.5}\) & 43.9\({\pm0.5}\) \\
\bottomrule
\end{tabular}
}
\label{tab:domain_label}
\end{table}

To evaluate the effectiveness of our pseudo-domain labeling strategy, we further experimented with log-variance (log-var) for extracting domain-specific features and K-Means clustering for clustering the extracted features to assign pseudo-domain labels.
The feature for log-variance is extracted from the first and second layers of the three-depth ConvNet.
Additionally, we compare the results with the random pseudo-domain labeling and actual domain labels for the available datasets, PACS and VLCS.
The random pseudo-domain labeling is done by assigning a pseudo-domain label for each synthetic image not pixel-wise as done in DAM.
The experiments are conducted using DM+DAM across three datasets under 1 and 10 IPC and the results are shown in Table~\ref{tab:domain_label}.
Across all datasets and IPC configurations, FFT-based feature extraction consistently outperforms log-variance, regardless of the clustering strategy applied.
Notably, the combination of FFT and Mean-Sort achieves the highest performance and even surpasses the use of actual domain labels.
\setlength{\intextsep}{0pt}
\begin{wrapfigure}{r}{0.6\textwidth}
    \centering
    \includegraphics[width=0.58\textwidth]{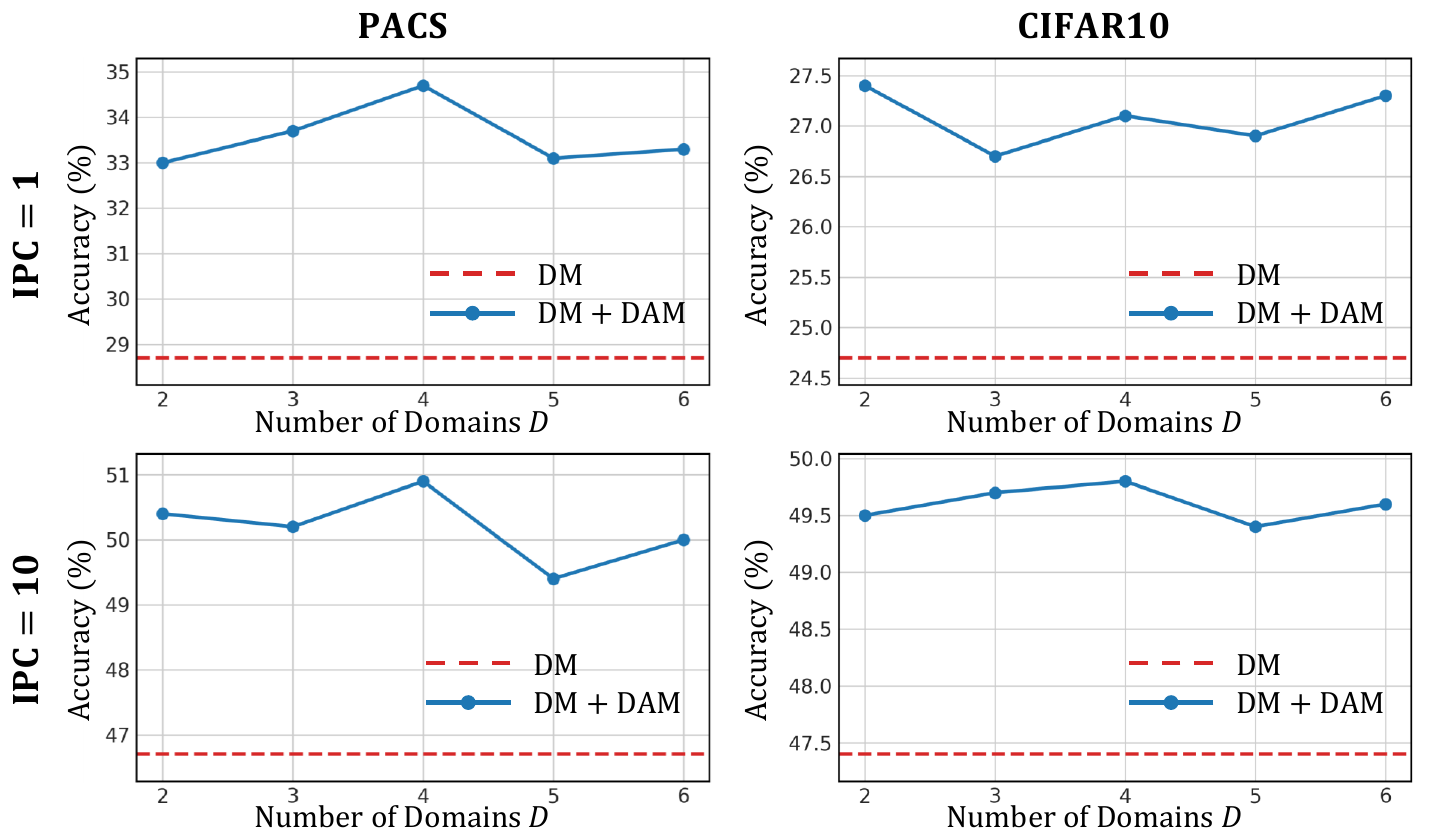}
    \caption{Experiment with a varying number of domains $D$ on CIFAR-10 and PACS dataset under 1 and 10 IPC with DM and DM+DAM.}
    \label{fig:numberofdomains}
\end{wrapfigure}In contrast, random pseudo-domain labeling yields the lowest performance among all variants, though it still performs better than DM without DAM, highlighting the value of incorporating even weak domain information.

\vspace{-2ex}

\paragraph{Effect of the Number of Pseudo Domains}
We analyze the impact of the number of pseudo domains \(D\) on the performance of DM+DAM across CIFAR-10 and PACS under both 1 and 10 IPC.
As shown in Figure~\ref{fig:numberofdomains}, the dashed red line indicates the performance of the baseline DM method, while the solid blue curve shows the performance of DM+DAM as \(D\) varies.
In all settings, DM+DAM consistently outperforms DM, demonstrating the effectiveness of incorporating domain-specific information during condensation.
Notably, for PACS, which has four explicit annotated domains, the best performance is observed when \(D = 4\) in both IPC settings, suggesting that the number of pseudo domains aligned with the true domain numbers is especially beneficial.
On the other hand, the optimal number of pseudo domains in CIFAR-10 varies across settings, indicating that the best partitioning may depend on the nature of the dataset and the number of images per class, however, we emphasize that even with random number of domains \(D\), we achieve better performance than basline methods (i.e. w/o DAM).
\vspace{-1ex}
\section{Conclusion}
\label{sec:conclusion}
\vspace{-1ex}

In this work, we introduce Multi-Domain Dataset Condensation (MDDC), the first framework to explicitly tackle dataset condensation under multi-domain settings.
To tackle this setting, we propose the Domain-Aware Module (DAM), a plug-and-play component that embeds domain-specific information into synthetic data.
Unlike prior methods that focus solely on class preservation, DAM leverages domain masking and FFT-based pseudo-domain labeling to preserve domain diversity, improving both in-domain performance and out-of-domain generalization.
Extensive experiments across various IPC settings, datasets, and architectures confirm the effectiveness of our approach.
While DAM introduces additional components such as domain mask optimization and pseudo-label assignment during training, it establishes a strong foundation for future research in multi-domain dataset condensation.

\vspace{-2ex}

\paragraph{Limitation}
DAM introduces additional parameters and a domain-specific loss, which increase condensation time and memory usage.
On the PACS dataset under 1 IPC, our method roughly doubles the condensation time due to the added domain classifier, while memory usage increases by only 0.1GB.
Since condensation is a one-time process and its purpose is to reduce downstream training time, this overhead is typically negligible in practice.
Detailed measurements are provided in the supplementary material~\ref{supple:computational_cost}.

\newpage
{\small
\bibliographystyle{IEEEtran}
\bibliography{egbib}
}

\newpage

\appendix
\renewcommand{\thetable}{\Alph{table}}
\setcounter{table}{0}

\section{Details regarding \(\Theta\) and \(\Theta'\)}
\label{supple:theta_details}

We clarify the roles of two sets of model parameters in our framework: \(\Theta\) and \(\Theta'\).
Both parameter sets correspond to models with the same architecture but serve different purposes and operate on different types of data batches.

The need for \(\Theta'\) arises from the fact that domain-aware loss must be computed over batches grouped by pseudo-domain labels (e.g., derived via FFT-based clustering), which differs from the class-wise batches typically used in condensation methods.
Moreover, in methods such as DC and MTT, where \(\Theta\) is either actively updated or pretrained for a specific matching loss, reusing the same parameter set for domain-aware supervision is unsuitable.
Thus, \(\Theta'\) is introduced to decouple domain-specific learning from class-based learning during the condensation process.

\begin{itemize}
    \item \textbf{{DC~\cite{dc_2021}}}: \(\Theta\) is randomly initialized and updated through bi-level optimization using class-wise batches. Because \(\Theta\) is trained throughout the condensation process, a separate parameter set \(\Theta'\) is introduced and trained on domain-grouped batches to compute the domain-aware loss independently.

    \item \textbf{DM~\cite{dm_2023}}: \(\Theta\) is randomly initialized but remains fixed throughout the condensation process. Since the parameters are not updated, the same \(\Theta\) can be reused to compute the domain-aware loss, and an explicit \(\Theta'\) is not required, even though domain-grouped batches are still used for the loss computation.

    \item \textbf{MTT~\cite{mtt_2022}}: \(\Theta\) is pretrained on real data and used to guide condensation via stored training trajectories. To preserve this role, a separate parameter set \(\Theta'\) is trained independently using \textbf{pseudo-domain labels} on real data prior to condensation, in a manner similar to the pretraining of \(\Theta\).
\end{itemize}

Across all methods incorporating domain-aware learning, batches used with \(\Theta'\) are consistently organized by pseudo-domain labels.
Whether a distinct \(\Theta'\) is needed depends on whether \(\Theta\) is trained, before or during the condensation process.

\section{Initializing synthetic data with real images}
\label{supple:real_initial}
\begin{table}[h]
\centering
\small
\caption{Performance comparison on CIFAR-10 and PACS datasets under 1 and 10 Image Per Class (IPC) settings. The experiment was done with \textbf{real} initializing the synthetic data following prior methods.}
\label{tab:supple_real_init}
\begin{tabularx}{\textwidth}{lYYYY}
\toprule
Dataset     & \multicolumn{2}{c}{CIFAR-10}                         & \multicolumn{2}{c}{PACS}                              \\
\cmidrule(lr){1-1} \cmidrule(lr){2-3} \cmidrule(lr){4-5}
Img/Cls     & 1                         & 10                        & 1                         & 10                        \\
Ratio (\%)  & 0.02                      & 0.2                       & 0.08                      & 0.8                       \\
\midrule
DC          & 28.2\(_{\pm0.6}\)           & 44.7\(_{\pm0.5}\)           & 35.9\(_{\pm1.1}\)           & 47.7\(_{\pm1.1}\)           \\
DC + DAM    & \textbf{29.0\(_{\pm0.4}\)}  & \textbf{45.2\(_{\pm0.3}\)}           & \textbf{37.8\(_{\pm0.7}\)}           & \textbf{48.7\(_{\pm0.2}\)}           \\
DM          & 25.7\(_{\pm0.6}\)           & 49.1\(_{\pm0.2}\)           & 32.0\(_{\pm1.9}\)           & 50.0\(_{\pm0.9}\)           \\
DM + DAM    & \textbf{26.8\(_{\pm0.3}\)}  & \textbf{50.0\(_{\pm0.3}\)}  & \textbf{32.7\(_{\pm1.2}\)}  & \textbf{50.9\(_{\pm0.6}\)}  \\
MTT         & 45.4\(_{\pm0.2}\)           & 65.3\(_{\pm0.4}\)           & 44.3\(_{\pm1.8}\)           & 51.4\(_{\pm1.2}\)           \\
MTT + DAM   & \textbf{45.6\(_{\pm0.3}\)}  & \textbf{65.5\(_{\pm0.2}\)}  & \textbf{44.4\(_{\pm1.6}\)}  & \textbf{53.5\(_{\pm1.3}\)}  \\
\bottomrule
\end{tabularx}
\end{table}

In the main manuscript, all experiments initialize synthetic data using \textbf{Gaussian noise}, which better aligns with the privacy-preserving goals of dataset condensation.
However, to demonstrate that our proposed method works even under alternative initializations, we conduct additional experiments where synthetic data is initialized with real images, selecting a random image from the corresponding class in the real dataset, following prior works.
These results, shown in Table~\ref{tab:supple_real_init}, represent averages over 10 runs, consistent with our main evaluation protocol.

While all methods benefit from real initialization, as expected due to the additional structure provided at the start, the performance gains from DAM persist, underscoring its robustness.
Notably, the relative improvement from DAM remains more pronounced in multi-domain settings like PACS, where domain shift presents a bigger challenge.
In contrast, single-domain datasets such as CIFAR-10 exhibit smaller domain-induced variability, which partially reduces the benefits of DAM when real images are used as initialization.

\section{Hyperparameter sweep}
\label{supple:hyperparameter_sweep}
\subsection{Domain mask initialization \(\textbf{z}_m\)}
\begin{table}[h]
\centering
\small
\caption{Effect of varying the domain mask initialization on CIFAR-10 and PACS datasets under IPC 1 and IPC 10.  All results are the average of 10 runs and reported as mean \(\pm\) standard deviation. The \colorbox{gray!20}{gray background setting} is the setting equal to the results in the main manuscript. The highest is \textbf{bolded} and the second highest is \underline{underlined}.}
\label{tab:supple_mask_init}
\begin{tabularx}{\textwidth}{llYYYY}
\toprule
\multicolumn{2}{c}{Dataset (\(\rightarrow\))} & \multicolumn{2}{c}{CIFAR-10} & \multicolumn{2}{c}{PACS} \\
\cmidrule(lr){1-2} \cmidrule(lr){3-4} \cmidrule(lr){5-6}
\multicolumn{2}{c}{Img/Cls (\(\rightarrow\))} & 1 & 10 & 1 & 10 \\
\multicolumn{2}{c}{Ratio (\%) (\(\rightarrow\))} & 0.02 & 0.2 & 0.08 & 0.8 \\
\midrule
Method (\(\downarrow\)) & Initial Value (\(\downarrow\)) & \\
\midrule
\multirow{5}{*}{DC + DAM} & 0.1 & 28.5\(_{\pm0.3}\) & \underline{45.3\(_{\pm0.4}\)} & 38.5\(_{\pm0.9}\) & \underline{48.8\(_{\pm0.7}\)} \\
& 0.05 & \underline{28.8\(_{\pm0.3}\)} & 45.2\(_{\pm0.4}\) & \underline{38.8\(_{\pm0.9}\)} & 47.9\(_{\pm0.7}\) \\
& \cellcolor{gray!20}0.01 & \cellcolor{gray!20} \textbf{29.0\(_{\pm0.5}\)} & \cellcolor{gray!20} \textbf{45.4\(_{\pm0.3}\)} & \cellcolor{gray!20} \underline{38.8\(_{\pm0.7}\)} & \cellcolor{gray!20} 48.3\(_{\pm0.5}\) \\
& 0.005 & 28.7\(_{\pm0.5}\) & 45.2\(_{\pm0.2}\) & 38.6\(_{\pm0.4}\) & \textbf{49.0\(_{\pm0.7}\)} \\
& 0.001 & 28.7\(_{\pm0.6}\) & 45.1\(_{\pm0.4}\) & \textbf{39.4\(_{\pm0.6}\)} & 47.9\(_{\pm0.6}\) \\

\midrule
\multirow{5}{*}{DM + DAM} & 0.1 & \textbf{27.2\(_{\pm0.3}\)} & 49.7\(_{\pm0.3}\) & \underline{34.2\(_{\pm1.6}\)} & 50.1\(_{\pm0.5}\) \\
& 0.05 & 26.6\(_{\pm0.3}\) & \textbf{49.9\(_{\pm0.4}\)} & \underline{34.2\(_{\pm0.5}\)} & 49.7\(_{\pm0.6}\)\\
& \cellcolor{gray!20}0.01 & \cellcolor{gray!20} \underline{27.1\(_{\pm0.3}\)} & \cellcolor{gray!20} \underline{49.8\(_{\pm0.5}\)} & \cellcolor{gray!20} \textbf{34.7\(_{\pm0.8}\)} & \cellcolor{gray!20} \textbf{50.9\(_{\pm0.4}\)} \\
& 0.005 & 26.3\(_{\pm0.3}\) & 48.9\(_{\pm0.2}\) & 34.2\(_{\pm1.0}\) & 49.2\(_{\pm0.7}\) \\
& 0.001 & 26.2\(_{\pm0.4}\) & 48.3\(_{\pm0.2}\) & 33.8\(_{\pm1.2}\) & \underline{50.8\(_{\pm0.4}\)} \\
\midrule
\multirow{5}{*}{MTT + DAM} & 0.1 & \underline{46.4\(_{\pm0.4}\)} & - & 43.0\(_{\pm1.2}\) & -  \\
& 0.05 & 41.1\(_{\pm0.6}\) & - & 43.1\(_{\pm1.0}\) & -  \\
& \cellcolor{gray!20}0.01 & \cellcolor{gray!20} \textbf{46.8\(_{\pm0.4}\)} & \cellcolor{gray!20} \textbf{57.9\(_{\pm0.4}\)} & \cellcolor{gray!20} \textbf{46.6\(_{\pm1.3}\)} & \cellcolor{gray!20} \textbf{50.6\(_{\pm0.6}\)} \\
& 0.005 & 41.2\(_{\pm0.8}\) & - & \underline{43.3\(_{\pm1.2}\)} & -  \\
& 0.001 & 41.9\(_{\pm0.5}\) & - & 40.7\(_{\pm1.2}\) & -  \\
\bottomrule
\end{tabularx}
\end{table}

The domain mask initialization value (\(z_m\)) controls the initial scale of the softmax-normalized spatial masks applied in the Domain-Aware Module (DAM).
A smaller \(z_m\) leads to nearly uniform domain weights at the beginning of training, allowing all domain masks to contribute equally.
In contrast, a larger \(z_m\) produces more confident, peaked softmax outputs early on, encouraging the model to assign higher importance to specific domains from the start.
To understand the effect of this initialization, we conduct a sweep across a range of \(z_m\) values and showcase the result in Table~\ref{tab:supple_mask_init}.

We find that the setting used in the main manuscript (\(z_m = 0.01\), gray-highlighted) consistently results in first- or second-best performance across all methods and datasets.
Crucially, even under different initializations, methods with DAM uniformly outperform their respective baselines without DAM, indicating strong robustness.
While MTT + DAM shows slightly more variation across \(z_m\) values compared to other methods, it still maintains a clear performance margin over MTT without DAM.
Due to observed instability at 1 IPC, we omit MTT + DAM results for IPC 10 in this ablation.

Overall, these results confirm that \(z_m = 0.01\) is a reliable and effective choice, and that DAM consistently enhances performance across settings.

\subsection{Domain embedding weight \(\lambda\)}
\begin{table}[h]
\centering
\small
\caption{Effect of varying the embedding weight on CIFAR-10 and PACS datasets under IPC 1 and IPC 10.  All results are the average of 10 runs and reported as mean \(\pm\) standard deviation. The \colorbox{gray!20}{gray background setting} is the setting equal to the results in the main manuscript. The highest is \textbf{bolded} and the second highest is \underline{underlined}}
\label{tab:supple_embedding_weight}
\begin{tabularx}{\textwidth}{llYYYY}
\toprule
\multicolumn{2}{c}{Dataset (\(\rightarrow\))} & \multicolumn{2}{c}{CIFAR-10} & \multicolumn{2}{c}{PACS} \\
\cmidrule(lr){1-2} \cmidrule(lr){3-4} \cmidrule(lr){5-6}
\multicolumn{2}{c}{Img/Cls (\(\rightarrow\))} & 1 & 10 & 1 & 10 \\
\multicolumn{2}{c}{Ratio (\%) (\(\rightarrow\))} & 0.02 & 0.2 & 0.08 & 0.8 \\
\midrule
Method (\(\downarrow\)) & \(\lambda\) (\(\downarrow\)) & \\
\midrule
\multirow{5}{*}{DC + DAM} & \cellcolor{gray!20}0.1 & \cellcolor{gray!20} \textbf{29.0\(_{\pm0.5}\)} & \cellcolor{gray!20} 45.4\(_{\pm0.3}\) & \cellcolor{gray!20} \underline{38.8\(_{\pm0.7}\)} & \cellcolor{gray!20} \textbf{48.3\(_{\pm0.5}\)} \\
& 0.05 & \textbf{29.0\(_{\pm0.2}\)} & 44.9\(_{\pm0.3}\) & \textbf{38.9\(_{\pm0.7}\)} & 47.7\(_{\pm0.3}\) \\
& 0.01 & \underline{28.9\(_{\pm0.6}\)} & \textbf{45.6\(_{\pm0.2}\)} & \underline{38.8\(_{\pm0.4}\)} & \underline{48.0\(_{\pm0.5}\)} \\
& 0.005 & 28.8\(_{\pm0.4}\) & 45.0\(_{\pm0.2}\) & 38.3\(_{\pm0.5}\) & 47.7\(_{\pm0.5}\) \\
& 0.001 & \underline{28.9\(_{\pm0.5}\)} & \underline{45.7\(_{\pm0.3}\)} & 38.6\(_{\pm0.4}\) & \underline{48.0\(_{\pm0.4}\)} \\
\midrule
\multirow{5}{*}{DM + DAM} & \cellcolor{gray!20}0.1 & \cellcolor{gray!20} \underline{27.1\(_{\pm0.3}\)} & \cellcolor{gray!20} \textbf{49.8\(_{\pm0.5}\)} & \cellcolor{gray!20} \textbf{34.7\(_{\pm0.8}\)} & \cellcolor{gray!20} \textbf{50.9\(_{\pm0.4}\)} \\
& 0.05 & \textbf{27.2\(_{\pm0.5}\)} & 48.9\(_{\pm0.4}\) & \underline{34.3\(_{\pm1.3}\)} & 49.3\(_{\pm0.4}\) \\
& 0.01 & 26.5\(_{\pm0.2}\) & 49.0\(_{\pm0.4}\) & \underline{34.3\(_{\pm0.6}\)} & 50.4\(_{\pm0.7}\) \\
& 0.005 & 26.7\(_{\pm0.6}\) & 48.9\(_{\pm0.3}\) & \underline{34.3\(_{\pm0.7}\)} & 50.0\(_{\pm0.6}\)\\
& 0.001 & 26.8\(_{\pm0.4}\) & \underline{49.5\(_{\pm0.4}\)} & 33.3\(_{\pm0.7}\) & \underline{50.5\(_{\pm0.8}\)} \\
\midrule
\multirow{5}{*}{MTT + DAM} & 0.1 & 46.1\(_{\pm1.3}\) & - & 43.3\(_{\pm0.8}\) & - \\
& 0.05 & 46.5\(_{\pm0.8}\) & - & 43.9\(_{\pm0.5}\) & - \\
& \cellcolor{gray!20}0.01 & \cellcolor{gray!20} \textbf{46.8\(_{\pm0.4}\)} & \cellcolor{gray!20} \textbf{57.9\(_{\pm0.4}\)} & \cellcolor{gray!20} \textbf{46.6\(_{\pm0.9}\)} & \cellcolor{gray!20} \textbf{50.6\(_{\pm0.6}\)} \\
& 0.005 & 46.0\(_{\pm0.5}\) & - & 46.0\(_{\pm0.7}\) & - \\
& 0.001 & \underline{46.7\(_{\pm0.8}\)} & - & \underline{46.5\(_{\pm1.0}\)} & - \\
\bottomrule\end{tabularx}
\end{table}

We study the effect of varying the domain embedding weight \(\lambda\), which balances the class loss and domain-aware loss in DAM.
A smaller \(\lambda\) reduces the influence of domain-specific learning, while a larger value encourages the model to attend more strongly to domain variations during condensation.

As shown in Table~\ref{tab:supple_embedding_weight}, performance remains strong across a wide range of \(\lambda\) values, showing that the method is not overly sensitive to this hyperparameter.
The setting used in the main manuscript (\(\lambda = 0.1\), gray-highlighted) consistently achieves the best or second-best performance across datasets and methods.
This confirms that \(\lambda = 0.1\) is a reliable default, and that DAM provides robust improvements without requiring precise tuning.

We do not explore values of \(\lambda\) greater than 0.1, as assigning excessive weight to domain supervision risks overshadowing class-discriminative learning.
As emphasized in the main manuscript, DAM is designed to enrich class information with domain cues, not to compete with it.

As with the previous sweep, we omit MTT + DAM results for IPC 10 due to instability observed under 1 IPC setting.

\subsection{Temperature \(\tau\)}
\begin{table}[h]
\centering
\small
\caption{Effect of varying the temperature \(\tau\) on CIFAR-10 and PACS datasets under IPC 1 and IPC 10.  All results are the average of 10 runs and reported as mean \(\pm\) standard deviation. The \colorbox{gray!20}{gray background setting} is the setting equal to the results in the main manuscript. The highest is \textbf{bolded} and the second highest is \underline{underlined}}
\label{tab:supple_tau}
\begin{tabularx}{\textwidth}{llYYYY}
\toprule
\multicolumn{2}{c}{Dataset (\(\rightarrow\))} & \multicolumn{2}{c}{CIFAR-10} & \multicolumn{2}{c}{PACS} \\
\cmidrule(lr){1-2} \cmidrule(lr){3-4} \cmidrule(lr){5-6}
\multicolumn{2}{c}{Img/Cls (\(\rightarrow\))} & 1 & 10 & 1 & 10 \\
\multicolumn{2}{c}{Ratio (\%) (\(\rightarrow\))} & 0.02 & 0.2 & 0.08 & 0.8 \\
\midrule
Method (\(\downarrow\)) & \(\tau\) (\(\downarrow\)) &  \\
\midrule
\multirow{3}{*}{DC + DAM} & \cellcolor{gray!20}0.1 & \cellcolor{gray!20} \textbf{29.0\(_{\pm0.5}\)} & \cellcolor{gray!20} \textbf{45.4\(_{\pm0.3}\)} & \cellcolor{gray!20} \textbf{38.8\(_{\pm0.7}\)} & \cellcolor{gray!20} \underline{48.3\(_{\pm0.5}\)} \\
& 1 & 28.5\(_{\pm0.4}\) & \underline{45.3\(_{\pm0.3}\)} & \textbf{38.8\(_{\pm0.7}\)} & \textbf{48.7\(_{\pm0.5}\)} \\
& 5 & \underline{28.7\(_{\pm0.4}\)} & 45.1\(_{\pm0.3}\) & \textbf{38.8\(_{\pm0.5}\)} & \textbf{48.7\(_{\pm0.5}\)} \\
\midrule
\multirow{3}{*}{DM + DAM} & \cellcolor{gray!20}0.1 & \cellcolor{gray!20} \textbf{27.1\(_{\pm0.3}\)} & \cellcolor{gray!20} \textbf{49.8\(_{\pm0.5}\)} & \cellcolor{gray!20} \textbf{34.7\(_{\pm1.1}\)} & \cellcolor{gray!20} \textbf{50.9\(_{\pm0.4}\)} \\
& 1 & \underline{26.9\(_{\pm0.5}\)} & \underline{49.1\(_{\pm0.3}\)} & 33.1\(_{\pm1.0}\) & 49.8\(_{\pm0.5}\) \\
& 5 & 26.8\(_{\pm0.4}\) & \underline{49.1\(_{\pm0.4}\)} & \underline{33.3\(_{\pm1.0}\)} & \underline{50.4\(_{\pm0.5}\)} \\
\midrule
\multirow{3}{*}{MTT + DAM} & \cellcolor{gray!20}0.1 & \cellcolor{gray!20} \textbf{46.8\(_{\pm0.4}\)} & \cellcolor{gray!20} \textbf{57.9\(_{\pm0.4}\)} & \cellcolor{gray!20} 46.6\(_{\pm0.9}\) & \cellcolor{gray!20} \textbf{50.6\(_{\pm0.6}\)} \\
& 1 & \underline{42.4\(_{\pm0.6}\)} & - & \underline{46.9\(_{\pm0.4}\)} & - \\
& 5 & 42.0\(_{\pm0.7}\) & - & \textbf{48.5\(_{\pm0.8}\)} & - \\
\bottomrule
\end{tabularx}
\end{table}

We ablate the softmax temperature \(\tau\) in DAM, which controls the sharpness of domain assignment.
A lower \(\tau\) (e.g., 0.1) enforces peaked domain masks, while higher values (e.g., 1 or 5) blend domain cues more evenly.

As demonstrated in Table~\ref{tab:supple_tau}, experiments with \(\tau = 1\) and \(\tau = 5\) show similar results, whereas the more discriminative setting \(\tau = 0.1\) yields a vivid improvement in most configurations.
As with the previous sweep, we omit MTT + DAM results for IPC 10 due to instability observed under 1 IPC setting.

\section{Additional dataset for multi-domain setting}
\begin{table}[h]
\setlength{\tabcolsep}{4pt}
\centering
\small
\caption{Comparison of DomainNet under IPC 1.  All results are the average of 3 runs and reported as mean \(\pm\) standard deviation.}
\label{tab:supple_domainnet}
\begin{tabularx}{\textwidth}{lY}
\toprule
Dataset & DomainNet~\cite{domainnet} \\
\midrule
Img/Cls & 1 \\
Ratio (\%) & 0.06 \\
\midrule
DM          & 3.44\(_{\pm0.03}\) \\
DM + DAM    & \textbf{3.52\(_{\pm0.03}\)} \\
\bottomrule
\end{tabularx}
\end{table}
To evaluate the scalability of our approach on a larger multi-domain dataset, we conduct experiments on DomainNet~\cite{domainnet}, a benchmark dataset comprising 345 classes across six distinct domains: Clipart, Infograph, Painting, Quickdraw, Real, and Sketch.
The total dataset contains approximately 586,575 images, with the number of samples per domain ranging from 48,837 to 175,327, making it one of the largest and most diverse domain generalization datasets.

Due to the high computational demand of such a large-scale dataset, we perform the evaluation under the 1 Image Per Class (IPC) setting, which corresponds to a 0.06\% data ratio.
The results are reported in Table~\ref{tab:supple_domainnet}.
While the overall performance is lower, owing to the dataset's complexity and extreme data compression, the incorporation of DAM still provides a measurable improvement over the baseline DM method, further demonstrating the robustness and scalability of our proposed approach.

\section{Computational cost}
\label{supple:computational_cost}
\begin{table}[t!]
\caption{Results with and without DAM on the prior methods on the \textbf{single-domain} setting. ``T.Image.'' denotes Tiny ImageNet dataset. The results are shown as peak GPU consumption - average training loop time.}
\label{tab:supple_consumption1}
\renewcommand{\arraystretch}{0.95}
\setlength{\tabcolsep}{4pt}
\centering
\small
\begin{tabularx}{\textwidth}{lYYYYYYY}
\toprule
\textbf{Dataset} & \multicolumn{3}{c}{\textbf{CIFAR-10}} & \multicolumn{3}{c}{\textbf{CIFAR-100}} & \multicolumn{1}{c}{\textbf{T.Image.}} \\
\cmidrule(lr){1-1} \cmidrule(lr){2-4} \cmidrule(lr){5-7} \cmidrule(lr){8-8}
Img/Cls & 1 & 10 & 50 & 1 & 10 & 50 & 1 \\
\midrule
\midrule
DC & \scriptsize{1GiB - 0.2s} & \scriptsize{1GiB - 11.1s} & \scriptsize{2GiB - 60.6s} & \scriptsize{2GiB - 1.8s} & \scriptsize{5GiB - 105s} & - & - \\
\textbf{DC + DAM} & \scriptsize{1GiB - 0.4s} & \scriptsize{2GiB - 13.4s} & \scriptsize{9GiB - 90.9s} & \scriptsize{2GiB - 2.4s} & \scriptsize{18GiB - 119s} & - & - \\
[0.1ex] \cdashline{1-8}[3pt/3pt] \\ [-1.8ex]
DM & \scriptsize{0.1GiB - 0.1s} & \scriptsize{1GiB - 0.1s} & \scriptsize{1GiB - 0.1s} & \scriptsize{1GiB - 0.7s} & \scriptsize{2GiB - 0.8s} & \scriptsize{8GiB - 0.8s} & \scriptsize{6GiB - 3.5s} \\
\textbf{DM + DAM} & \scriptsize{0.1GiB - 0.1s} & \scriptsize{1GiB - 0.2s} & \scriptsize{4GiB - 0.3s} & \scriptsize{1GiB - 1.2s} & \scriptsize{7GiB - 1.4s} & \scriptsize{36GiB - 2.5s} & \scriptsize{10GiB - 5.4s} \\
[0.1ex] \cdashline{1-8}[3pt/3pt] \\ [-1.8ex]
MTT & \scriptsize{1GiB - 2.2s} & \scriptsize{5GiB - 1.3s} & - & \scriptsize{5GiB - 2.4s} & - & - & \\
\textbf{MTT + DAM} & \scriptsize{1GiB - 3.3s} & \scriptsize{9GiB - 4.7s} & - & \scriptsize{9GiB - 4.5s} & - & - & \\
\bottomrule
\end{tabularx}
\end{table}

\begin{table}[t!]
\caption{Results with and without DAM on the prior methods on the \textbf{multi-domain} setting. The results are shown as peak GPU consumption - average training loop time.}
\label{tab:supple_consumption2}
\renewcommand{\arraystretch}{0.95}
\centering
\small
\begin{tabularx}{\textwidth}{lYYYYYYY}
\toprule
\textbf{Dataset} & \multicolumn{2}{c}{\textbf{PACS}} & \multicolumn{2}{c}{\textbf{VLCS}} & \multicolumn{2}{c}{\textbf{Office-Home}} \\
\cmidrule(lr){1-1} \cmidrule(lr){2-3} \cmidrule(lr){4-5} \cmidrule(lr){6-7}
Img/Cls & 1 & 10 & 1 & 10 & 1 & 10 \\
\midrule
\midrule
DC & \scriptsize{3GiB - 0.4s} & \scriptsize{4GiB - 26.6s} & \scriptsize{3GiB - 0.3s} & \scriptsize{3GiB - 18.9s} & \scriptsize{4GiB - 2.9s} & - \\
\textbf{DC + DAM} & \scriptsize{3GiB - 0.7s} & \scriptsize{5GiB - 31.4s} & \scriptsize{3GiB - 0.5s} & \scriptsize{4GiB - 22.9s} & \scriptsize{5GiB - 3.5s} & - \\
[0.1ex] \cdashline{1-7}[3pt/3pt] \\ [-1.8ex]
DM & \scriptsize{1GiB - 0.1s} & \scriptsize{1GiB - 0.2s} & \scriptsize{1GiB - 0.1s} & \scriptsize{1GiB - 0.1s} & \scriptsize{2GiB - 1s} & \scriptsize{5GiB - 1.1s}\\
\textbf{DM + DAM} & \scriptsize{1GiB - 0.2s} & \scriptsize{2GiB - 0.2s} & \scriptsize{1GiB - 0.2s} & \scriptsize{2GiB - 0.3s} & \scriptsize{2GiB - 1.2s} & \scriptsize{10GiB - 1.9s} \\
[0.1ex] \cdashline{1-7}[3pt/3pt] \\ [-1.8ex]
MTT & \scriptsize{1GiB - 2.5s} & \scriptsize{13GiB - 2.3s} & \scriptsize{1GiB - 2.8s}  & - & \scriptsize{12GiB - 3.0s}  & -\\
\textbf{MTT + DAM} & \scriptsize{3GiB - 4.0s} & \scriptsize{25GiB - 5.1s} & \scriptsize{2GiB - 4.1s} & - & \scriptsize{24GiB - 9.5s} & -\\
\bottomrule
\end{tabularx}
\end{table}

We report the GPU memory usage and per-iteration training time for with and without our method, DAM.
All experiments were conducted using an \textit{NVIDIA RTX A6000} GPU and an \textit{Intel Xeon Gold 6442Y} CPU.
The reported training time is the average duration of a single training loop measured over 10 iterations, taken after 10 warm-up iterations.
Peak GPU memory consumption is measured during the same window using PyTorch's memory profiling utilities (\texttt{torch.cuda.max\_memory\_allocated()}).

As shown in Table~\ref{tab:supple_consumption1} and Table~\ref{tab:supple_consumption2}, and also noted in the limitation, incorporating DAM introduces an overhead.
As we introduce the domain masks per image, GPU memory usage increases with images per class (IPC) and the number of dataset classes.
However, the training time doubles only in the low IPC and does not linearly grow with the IPC and the number of dataset classes.

For MTT, we observed a slightly different behavior.
GPU memory usage and training time were unstable across repeated runs, with noticeable fluctuations.
We attribute this instability to the overhead of loading and processing trajectory data within each training loop, which is unique to the MTT framework.
Due to this inconsistency, we report the highest observed GPU memory usage and training time across three repeated runs for each setting.

\section{Hyperparameter for MTT}
\begin{table}[t!]
\caption{Hyperparameters for DAM with MTT. ``T.Image.'' denotes Tiny ImageNet dataset and ``OH'' denote Office Home dataset.}
\label{tab:supple_hyperparam}
\renewcommand{\arraystretch}{0.95}
\setlength{\tabcolsep}{4pt}
\centering
\small
\begin{tabularx}{\textwidth}{YYYYYYYY}
\toprule
Dataset & IPC & Synthetic Steps & Exper Epochs & Max Start Epochs & Learning Rate Image & Learning Rate & Starting Synthetic Step Size \\
\midrule
\midrule
\multirow{2}{*}{CIFAR-10} & 1 & 50 & 2 & 2 & 100 & 10\(^{-7}\) & 10\(^{-2}\) \\
& 10 & 30 & 2 & 20 & 10\(^{5}\) & 10\(^{-6}\) & 10\(^{-2}\) \\
\midrule
\multirow{2}{*}{CIFAR-100} & 1 & 20 & 3 & 20 & 10\(^{3}\) & 10\(^{-5}\) & 10\(^{-2}\) \\
& 10 & 20 & 2 & 20 & 10\(^{3}\) & 10\(^{-5}\) & 10\(^{-2}\) \\
\midrule
T.Image. & 1 & 10 & 2 & 10 & 10\(^{4}\) & 10\(^{-4}\) & 10\(^{-2}\) \\
\midrule
\multirow{2}{*}{PACS} & 1 & 10 & 2 & 10 & 10\(^{4}\) & 10\(^{-5}\) & 10\(^{-2}\) \\
& 10 & 20 & 2 & 40 & 10\(^{4}\) & 10\(^{-6}\) & 10\(^{-2}\) \\
\midrule
\multirow{2}{*}{VLCS} & 1 & 10 & 2 & 10 & 10\(^{4}\) & 10\(^{-6}\) & 10\(^{-2}\) \\
& 10 & 20 & 2 & 40 & 10\(^{4}\) & 10\(^{-6}\) & 10\(^{-2}\) \\
\midrule
OH & 1 & 10 & 2 & 10 & 10\(^{4}\) & 10\(^{-4}\) & 10\(^{-2}\) \\
\bottomrule
\end{tabularx}
\end{table}

For DC and DM, we adopted the hyperparameters used in their respective original implementations.
For multi-domain datasets, we followed the same configuration as used for Tiny ImageNet.
In contrast, MTT required a separate hyperparameter search due to frequent occurrences of \texttt{NaN} losses during training when combined with DAM and \textbf{Gaussian noise} initialization. 
Table~\ref{tab:supple_hyperparam} lists the hyperparameters used for MTT with DAM across all datasets.

We initially started with the settings reported in the original MTT paper~\cite{mtt_2022}, and conducted minimal adjustments only when instability (e.g., \texttt{NaN} gradients or diverging loss) was observed.
We constrained the search to a narrow range around the original values, preferring stability over aggressive tuning.
It is important to note that these are not hyperparameters introduced by our method (DAM) but rather those that existed from the MTT pipeline.

\section{Qualitative results}
\begin{figure}[t!]
    \centering
    \includegraphics[width=\textwidth]{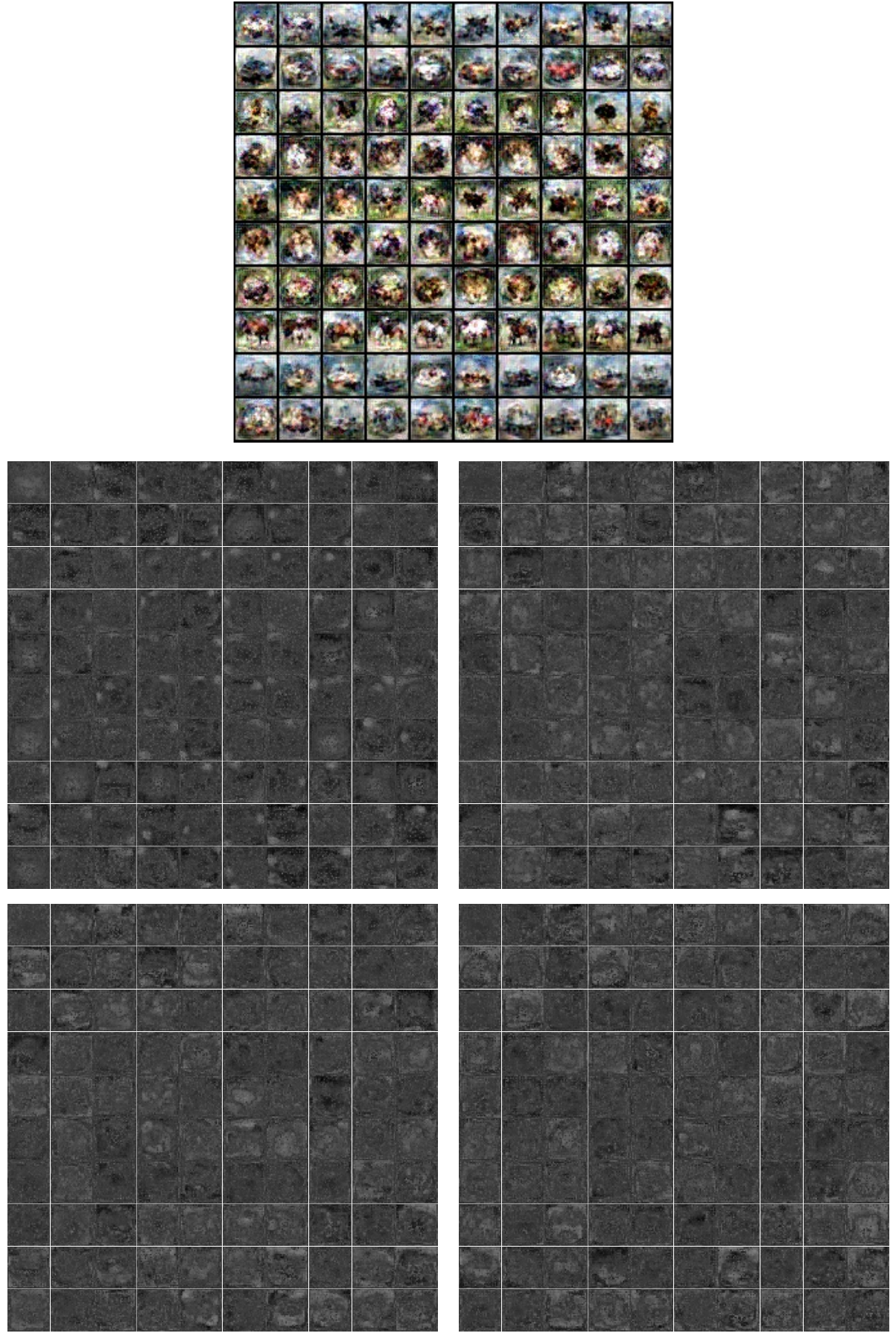}
    \caption{Visualization of the final output and domain masks in CIFAR-10 under 10 IPC setting. The shown images are condensed with DC+DAM.}
    \label{fig:supple_qualitative_results0}
\end{figure}

\begin{figure}[t!]
    \centering
    \includegraphics[width=\textwidth]{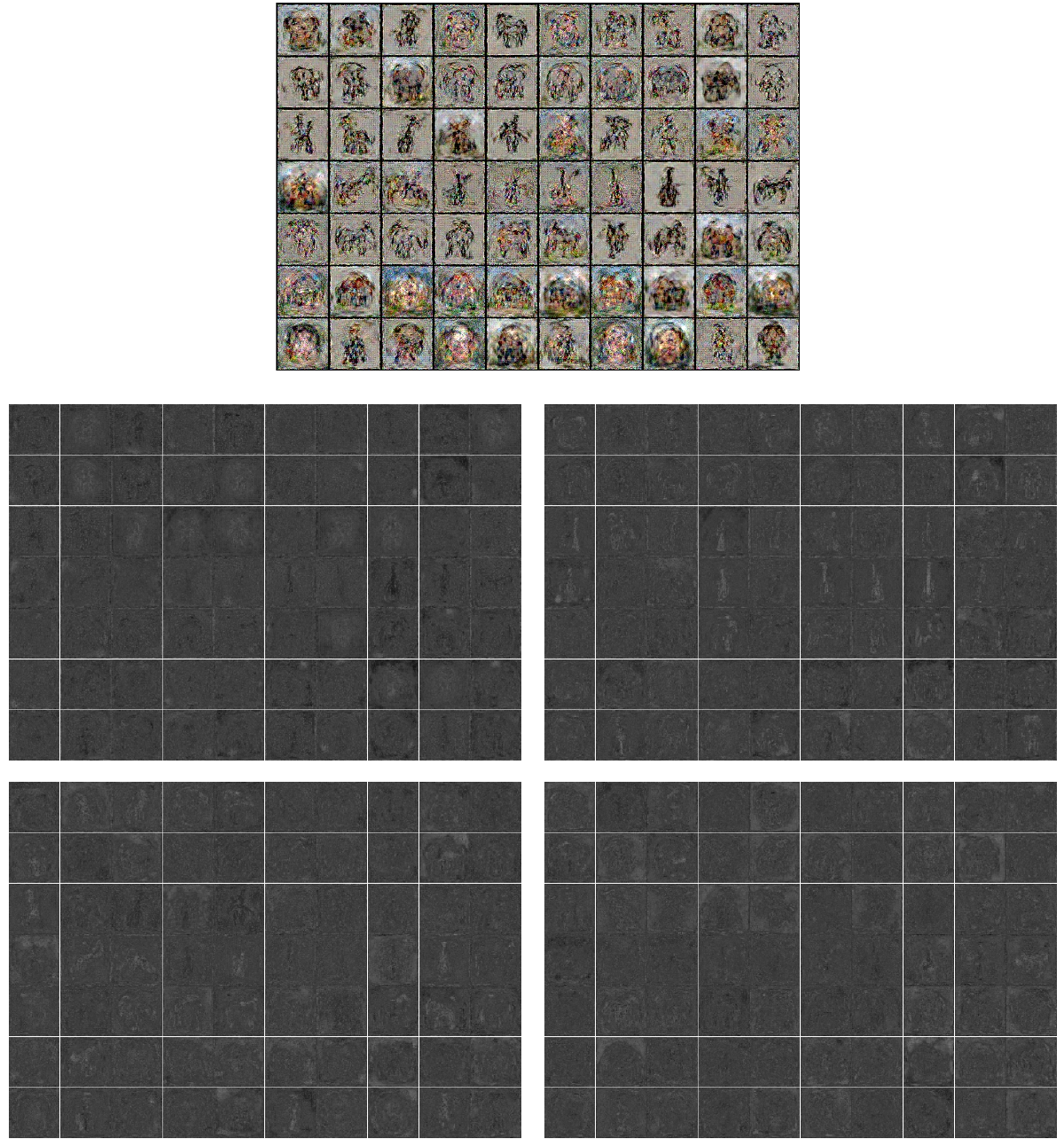}
    \caption{Visualization of the final output and domain masks in PACS under 10 IPC setting. The shown images are condensed with DC+DAM.}
    \label{fig:supple_qualitative_results1}
\end{figure}

\begin{figure}[t!]
    \centering
    \includegraphics[width=\textwidth]{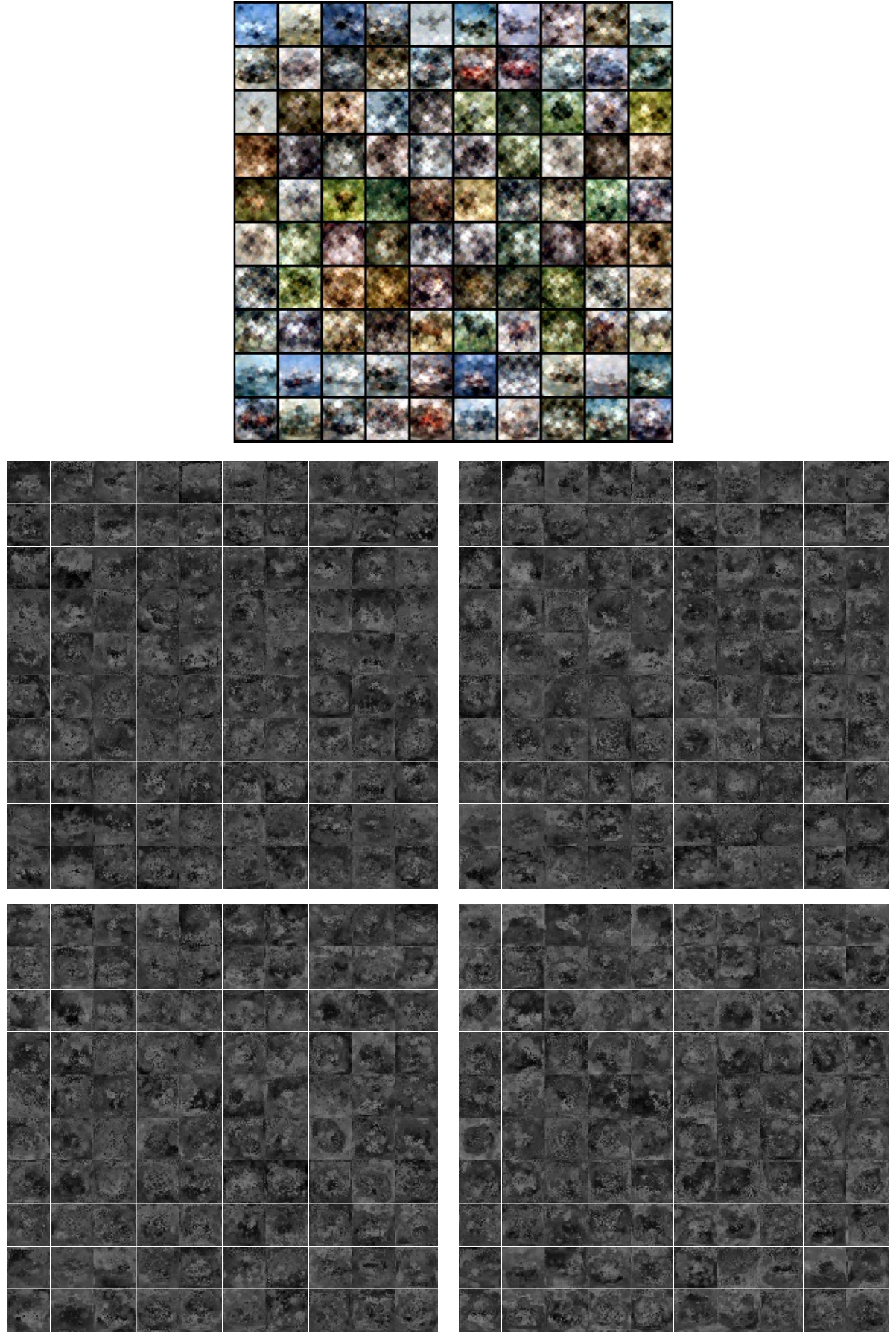}
    \caption{Visualization of the final output and domain masks in CIFAR-10 under 10 IPC setting. The shown images are condensed with DM+DAM.}
    \label{fig:supple_qualitative_results2}
\end{figure}

\begin{figure}[t!]
    \centering
    \includegraphics[width=\textwidth]{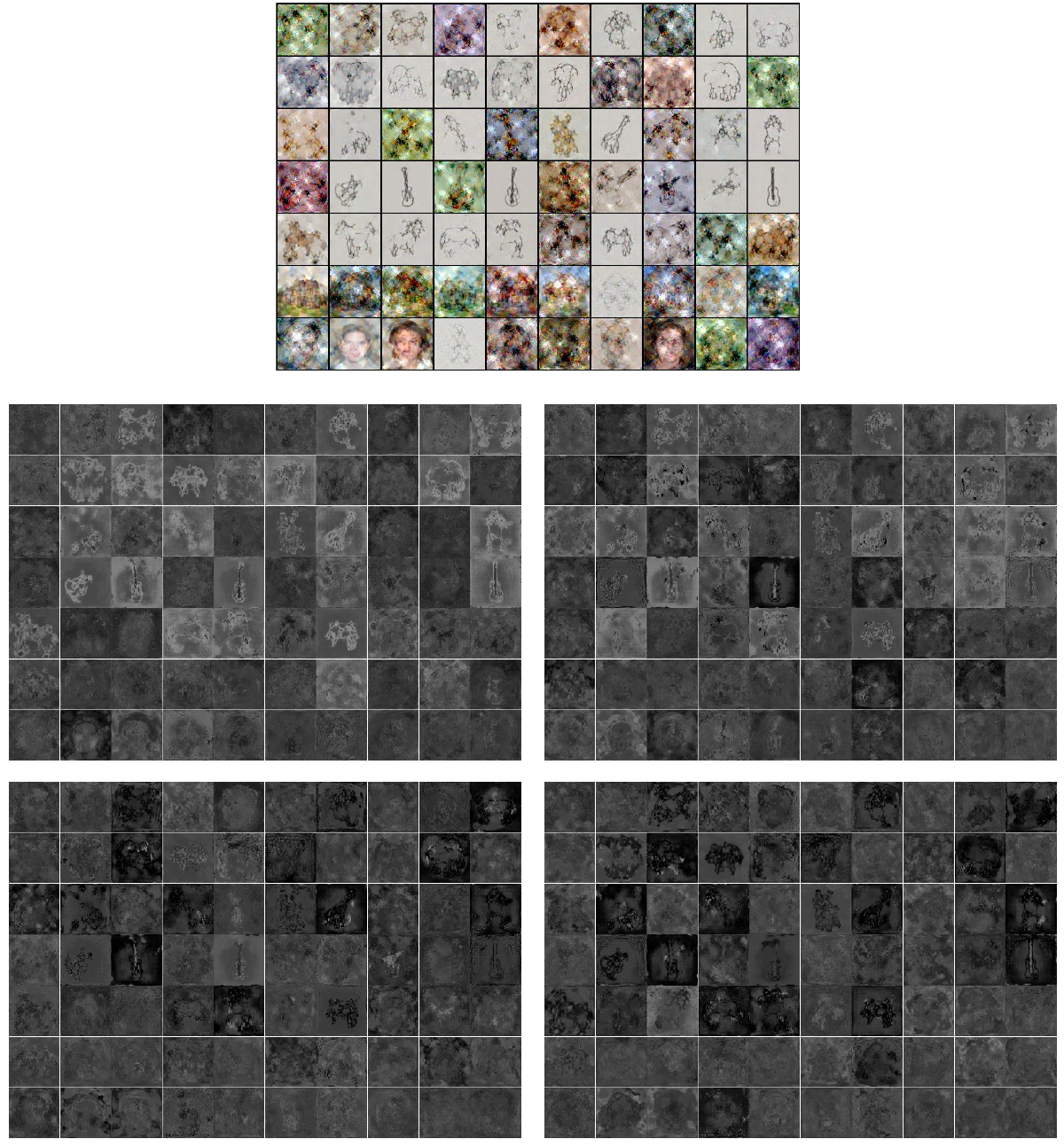}
    \caption{Visualization of the final output and domain masks in PACS under 10 IPC setting. The shown images are condensed with DM+DAM.}
    \label{fig:supple_qualitative_results3}
\end{figure}

We provide additional qualitative examples of the condensed synthetic images generated with DAM in Figure~\ref{fig:supple_qualitative_results0}, \ref{fig:supple_qualitative_results1}, \ref{fig:supple_qualitative_results2}, and \ref{fig:supple_qualitative_results3}.

All visualizations are obtained under IPC 10 using the CIFAR-10 and PACS datasets.
We present results based on DC and DM baselines, and visualize the synthetic images and domain mask after the final condensation step.

\end{document}